\documentclass[12pt,a4paper]{article}

\usepackage[a4paper,text={16.5cm,25.2cm},centering]{geometry}
\usepackage{lmodern}
\usepackage{amssymb,amsmath}
\usepackage{graphicx}
\usepackage{microtype}
\usepackage{hyperref}
\hypersetup
       {   pdfauthor = { Chris Rackauckas, Mike Innes, Yingbo Ma, \\ Jesse Bettencourt,
Lyndon White, Vaibhav Dixit },
           pdftitle={ DiffEqFlux.jl - A Julia Library for Neural Differential Equations },
           colorlinks=TRUE,
           linkcolor=black,
           citecolor=blue,
           urlcolor=blue
       }

\title{ DiffEqFlux.jl --- A Julia Library for\\ Neural Differential Equations }

\author{ Chris Rackauckas, Mike Innes, Yingbo Ma, \\ Jesse Bettencourt,
Lyndon White, Vaibhav Dixit }

\usepackage[T1]{fontenc}
\usepackage{textcomp}
\usepackage{upquote}
\usepackage{listings}
\usepackage{xcolor}
\newcommand{\HLJLk}[1]{\textcolor[RGB]{148,91,176}{\textbf{#1}}}

\newcommand{\HLJLkp}[1]{\textcolor[RGB]{148,91,176}{\textbf{#1}}}

\newcommand{\HLJLn}[1]{#1}

\newcommand{\HLJLnd}[1]{\textcolor[RGB]{214,102,97}{#1}}

\newcommand{\HLJLnf}[1]{\textcolor[RGB]{66,102,213}{#1}}

\newcommand{\HLJLs}[1]{\textcolor[RGB]{201,61,57}{#1}}

\newcommand{\HLJLnfB}[1]{\textcolor[RGB]{59,151,46}{#1}}

\newcommand{\HLJLni}[1]{\textcolor[RGB]{59,151,46}{#1}}

\newcommand{\HLJLoB}[1]{\textcolor[RGB]{102,102,102}{\textbf{#1}}}

\newcommand{\HLJLp}[1]{#1}

\newcommand{\HLJLcs}[1]{\textcolor[RGB]{153,153,119}{\textit{#1}}}

\begin{document}

\maketitle

\abstract{DiffEqFlux.jl is a library for fusing neural networks and differential equations.
In this work we describe differential equations from the viewpoint of data science and discuss the complementary nature between machine learning models and differential equations.
We demonstrate the ability to incorporate DifferentialEquations.jl-defined differential equation problems into a Flux-defined neural network, and vice versa.
The advantages of being able to use the entire DifferentialEquations.jl suite for this purpose is demonstrated by counter examples where simple integration strategies fail, but the sophisticated integration strategies provided by the DifferentialEquations.jl library succeed.
This is followed by a demonstration of delay differential equations and stochastic differential equations inside of neural networks. 
We show high-level functionality for defining neural ordinary differential equations (neural networks embedded into the differential equation) and describe the extra models in the Flux model zoo which includes neural stochastic differential equations.
We conclude by discussing the various adjoint methods used for backpropogation of the differential equation solvers.
DiffEqFlux.jl is an important contribution to the area, as it allows the full weight of the differential equation solvers developed from decades of research in the scientific computing field to be readily applied to the challenges posed by machine learning and data science.
}

\vspace{20px}
\fbox{%
\begin{minipage}{0.75\textwidth}
This paper is a republication of \href{https://julialang.org/blog/2019/01/fluxdiffeq}{this blog post} on the \href{https://github.com/JuliaDiffEq/DiffEqFlux.jl}{DiffEqFlux.jl package} into an academically standard format. The most up-to-date versions of the codes can be found in the \href{https://github.com/FluxML/model-zoo}{model-zoo}.
\end{minipage}
}
\vspace{20px}

In this blog post we will show you how to easily, efficiently, and robustly use differential equation (DiffEq) solvers with neural networks in Julia.

(Animations are omitted from the paper. Please see the \href{https://julialang.org/blog/2019/01/fluxdiffeq}{original blog post})

The \href{https://arxiv.org/abs/1806.07366}{Neural Ordinary Differential Equations} \cite{chen_neural_2018} paper has attracted significant attention even before it was awarded one of the Best Papers of NeurIPS 2018. The paper already gives many exciting results combining these two disparate fields, but this is only the beginning: neural networks and differential equations were born to be together. This blog post, a collaboration between authors of \href{https://github.com/FluxML/Flux.jl}{Flux} \cite{innes_flux:_2018}, \href{https://github.com/JuliaDiffEq/DifferentialEquations.jl}{DifferentialEquations.jl} \cite{christopher_rackauckas_differentialequations.jl_2017} and the Neural ODEs paper, will explain why, outline current and future directions for this work, and start to give a sense of what's possible with state-of-the-art tools.

The advantages of the Julia \href{https://github.com/JuliaDiffEq/DifferentialEquations.jl}{DifferentialEquations.jl} library for numerically solving differential equations have been \href{http://www.stochasticlifestyle.com/comparison-differential-equation-solver-suites-matlab-r-julia-python-c-fortran/}{discussed in detail in other posts}. Along with its \href{https://github.com/JuliaDiffEq/DiffEqBenchmarks.jl}{extensive benchmarking against classic Fortran methods}, it includes other modern features such as \href{http://www.stochasticlifestyle.com/solving-systems-stochastic-pdes-using-gpus-julia/}{GPU acceleration}, \href{http://docs.juliadiffeq.org/latest/features/monte_carlo.html}{distributed (multi-node) parallelism}, and \href{http://docs.juliadiffeq.org/latest/features/callback_functions.html}{sophisticated event handling}. Recently, these native Julia differential equation solvers have successfully been embedded into the \href{https://github.com/FluxML/Flux.jl/}{Flux} deep learning package, to allow the use of a full suite of highly tested and optimized DiffEq methods within neural networks. Using the new package \href{https://github.com/JuliaDiffEq/DiffEqFlux.jl/}{DiffEqFlux.jl}, we will show the reader how to easily add differential equation layers to neural networks using a range of differential equations models, including stiff ordinary differential equations, stochastic differential equations, delay differential equations, and hybrid (discontinuous) differential equations.

This is the first toolbox to combine a fully-featured differential equations solver library and neural networks seamlessly together. The blog post will also show why the flexibility of a full differential equation solver suite is necessary. With the ability to fuse neural networks with ODEs, SDEs, DAEs, DDEs, stiff equations, and different methods for adjoint sensitivity calculations, this is a large generalization of the neural ODEs work and will allow researchers to better explore the problem domain.

(Note: If you are interested in this work and are an undergraduate or graduate student, we have \href{https://julialang.org/soc/ideas-page}{Google Summer of Code projects available in this area}. This \href{https://developers.google.com/open-source/gsoc/help/student-stipends}{pays quite well over the summer}. Please join the \href{https://slackinvite.julialang.org/}{Julia Slack} and the \#jsoc channel to discuss in more detail.)

\section{What do differential equations have to do with machine learning?}
The first question someone not familiar with the field might ask is, why are differential equations important in this context? The simple answer is that a differential equation is a way to specify an arbitrary nonlinear transform by mathematically encoding prior structural assumptions.

Let's unpack that statement a bit. There are three common ways to define a nonlinear transform: direct modeling, machine learning, and differential equations. Directly writing down the nonlinear function only works if you know the exact functional form that relates the input to the output. However, in many cases, such exact relations are not known \emph{a priori}. So how do you do nonlinear modeling if you don't know the nonlinearity?

One way to address this is to use machine learning. In a typical machine learning problem, you are given some input $x$ and you want to predict an output $y$. This generation of a prediction $y$ from $x$ is a machine learning model (let's call it $ML$).  During training, we attempt to adjust the parameters of $ML$ so that it generates accurate predictions.  We can then use $ML$ for inference (i.e., produce $y$s for novel inputs $x$). This is just a nonlinear transformation $y=ML(x)$. The reason $ML$ is interesting is because its form is basic but adapts to the data itself. For example, a simple neural network (in design matrix form) with sigmoid activation functions is simply matrix multiplications followed by application of sigmoid functions. Specifically,  $ML(x)=\sigma(W_{3}\sigma(W_{2}\sigma(W_{1} x)))$ is a three-layer deep neural network, where $W=(W_1,W_2,W_3)$ are learnable parameters. You then choose $W$ such that $ML(x)=y$ reasonably fits the function you wanted it to fit. The theory and practice of machine learning confirms that this is a good way to learn nonlinearities. For example, the Universal Approximation Theorem states that, for enough layers or enough parameters (i.e. sufficiently large $W_{i}$ matrices), $ML(x)$ can approximate any nonlinear function sufficiently close (subject to common constraints).

So great, this always works! But it has some caveats, the main being that it has to learn everything about the nonlinear transform directly from the data. In many cases we do not know the full nonlinear equation, but we may know details about its structure. For example, the nonlinear function could be the population of rabbits in the forest, and we might know that their rate of births is dependent on the current population. Thus instead of starting from nothing, we may want to use this known \emph{a priori} relation and a set of parameters that defines it. For the rabbits, let's say that we want to learn
\begin{align}
\text{rabbits tomorrow} = \text{Model}(\text{rabbits today}).
\end{align}
In this case, we have prior knowledge of the rate of births being dependent on the current population. The way to mathematically state this structural assumption is via a differential equation. Here, what we are saying is that the birth rate of the rabbit population at a given time point increases when we have more rabbits. The simplest way of encoding that is
\begin{align}
\text{rabbits}'(t) = \alpha \text{rabbits}(t)
\end{align}
where $\alpha$ is some learnable constant. If you know your calculus, the solution here is exponential growth from the starting point with a growth rate $\alpha$: $\text{rabbits}(t_\text{start})e^{(\alpha t)}$. But notice that we didn't need to know the solution to the differential equation to validate the idea: we encoded the structure of the model and mathematics itself then outputs what the solution should be. Because of this, differential equations have been the tool of choice in most science. For example, physical laws tell you how electrical quantities emit forces (\href{https://en.wikipedia.org/wiki/Maxwell%27s_equations}{Maxwell's Equations}). These are essentially equations of how things change and thus "where things will be" is the solution to a differential equation. But in recent decades this application has gone much further, with fields like systems biology learning about cellular interactions by encoding known biological structures and mathematically enumerating our assumptions or in targeted drug dosage through PK/PD modelling in systems pharmacology.

So as our machine learning models grow and are hungry for larger and larger amounts of data, differential equations have become an attractive option for specifying nonlinearities in a learnable (via the parameters) but constrained form. They are essentially a way of incorporating prior domain-specific knowledge of the structural relations between the inputs and outputs. Given this way of looking at the two, both methods trade off advantages and disadvantages, making them complementary tools for modeling. It seems like a clear next step in scientific practice to start putting them together in new and exciting ways!

\section{What is the Neural Ordinary Differential Equation (ODE)?}
The neural ordinary differential equation is one of many ways to put these two subjects together. The simplest way of explaining it is that, instead of learning the nonlinear transformation directly, we wish to learn the structures of the nonlinear transformation. Thus instead of doing $y=ML(x)$, we put the machine learning model on the derivative, $y'(x) = ML(x)$, and now solve the ODE. Why would you ever do this? Well, one motivation is that defining the model in this way and then solving the ODE using the simplest and most error prone method, the Euler method, what you get is equivalent to a \href{https://arxiv.org/abs/1512.03385}{residual neural network}. The way the Euler method works is based on the fact that $y'(x) = \frac{dy}{dx}$, thus
\begin{align}
\Delta y = (y_\text{next} - y_\text{prev}) = \Delta x ML(x)
\end{align}
which implies that
\begin{align}
y_{i+1} = y_{i} + \Delta x ML(x_{i}).
\end{align}
This looks similar in structure to a ResNet, one of the most successful image processing models. The insight of the the Neural ODEs paper was that increasingly deep and powerful ResNet-like models effectively approximate a kind of "infinitely deep" model as each layer tends to zero. Rather than adding more layers, we can just model the differential equation directly and then solve it using a purpose-built ODE solver. Numerical ODE solvers are a science that goes all the way back to the first computers, and modern ones can adaptively choose step sizes $\Delta x$ and use high order approximations to dratically reduce the number of actual steps required. And as it turns out, this works well in practice, too.

\section{How do you solve an ODE?}
First, how do you numerically specify and solve an ODE? If you're new to solving ODEs, you may want to watch our \href{https://www.youtube.com/watch?v=KPEqYtEd-zY}{video tutorial on solving ODEs in Julia} and look through the \href{http://docs.juliadiffeq.org/latest/tutorials/ode_example.html}{ODE tutorial of the DifferentialEquations.jl documentation}. The idea is that you define an \texttt{ODEProblem} via a derivative equation \texttt{u'=f(u,p,t)}, and provide an initial condition \texttt{u0}, and a timespan \texttt{tspan} to solve over, and specify the parameters \texttt{p}.

For example, the \href{https://en.wikipedia.org/wiki/Lotka%E2%80%93Volterra_equations}{Lotka-Volterra equations describe the dynamics of the population of rabbits and wolves}. They can be written as:
\begin{align}
x^\prime &= \alpha x + \beta x y\\
y^\prime &= -\delta y + \gamma x y
\end{align}
and encoded in Julia like:
\begin{lstlisting}
(*@\HLJLk{using}@*) (*@\HLJLn{DifferentialEquations}@*)
(*@\HLJLk{function}@*) (*@\HLJLnf{lotka{\_}volterra}@*)(*@\HLJLp{(}@*)(*@\HLJLn{du}@*)(*@\HLJLp{,}@*)(*@\HLJLn{u}@*)(*@\HLJLp{,}@*)(*@\HLJLn{p}@*)(*@\HLJLp{,}@*)(*@\HLJLn{t}@*)(*@\HLJLp{)}@*)
  (*@\HLJLn{x}@*)(*@\HLJLp{,}@*) (*@\HLJLn{y}@*) (*@\HLJLoB{=}@*) (*@\HLJLn{u}@*)
  (*@\HLJLn{\ensuremath{\alpha}}@*)(*@\HLJLp{,}@*) (*@\HLJLn{\ensuremath{\beta}}@*)(*@\HLJLp{,}@*) (*@\HLJLn{\ensuremath{\delta}}@*)(*@\HLJLp{,}@*) (*@\HLJLn{\ensuremath{\gamma}}@*) (*@\HLJLoB{=}@*) (*@\HLJLn{p}@*)
  (*@\HLJLn{du}@*)(*@\HLJLp{[}@*)(*@\HLJLni{1}@*)(*@\HLJLp{]}@*) (*@\HLJLoB{=}@*) (*@\HLJLn{dx}@*) (*@\HLJLoB{=}@*) (*@\HLJLn{\ensuremath{\alpha}}@*)(*@\HLJLoB{*}@*)(*@\HLJLn{x}@*) (*@\HLJLoB{-}@*) (*@\HLJLn{\ensuremath{\beta}}@*)(*@\HLJLoB{*}@*)(*@\HLJLn{x}@*)(*@\HLJLoB{*}@*)(*@\HLJLn{y}@*)
  (*@\HLJLn{du}@*)(*@\HLJLp{[}@*)(*@\HLJLni{2}@*)(*@\HLJLp{]}@*) (*@\HLJLoB{=}@*) (*@\HLJLn{dy}@*) (*@\HLJLoB{=}@*) (*@\HLJLoB{-}@*)(*@\HLJLn{\ensuremath{\delta}}@*)(*@\HLJLoB{*}@*)(*@\HLJLn{y}@*) (*@\HLJLoB{+}@*) (*@\HLJLn{\ensuremath{\gamma}}@*)(*@\HLJLoB{*}@*)(*@\HLJLn{x}@*)(*@\HLJLoB{*}@*)(*@\HLJLn{y}@*)
(*@\HLJLk{end}@*)
(*@\HLJLn{u0}@*) (*@\HLJLoB{=}@*) (*@\HLJLp{[}@*)(*@\HLJLnfB{1.0}@*)(*@\HLJLp{,}@*)(*@\HLJLnfB{1.0}@*)(*@\HLJLp{]}@*)
(*@\HLJLn{tspan}@*) (*@\HLJLoB{=}@*) (*@\HLJLp{(}@*)(*@\HLJLnfB{0.0}@*)(*@\HLJLp{,}@*)(*@\HLJLnfB{10.0}@*)(*@\HLJLp{)}@*)
(*@\HLJLn{p}@*) (*@\HLJLoB{=}@*) (*@\HLJLp{[}@*)(*@\HLJLnfB{1.5}@*)(*@\HLJLp{,}@*)(*@\HLJLnfB{1.0}@*)(*@\HLJLp{,}@*)(*@\HLJLnfB{3.0}@*)(*@\HLJLp{,}@*)(*@\HLJLnfB{1.0}@*)(*@\HLJLp{]}@*)
(*@\HLJLn{prob}@*) (*@\HLJLoB{=}@*) (*@\HLJLnf{ODEProblem}@*)(*@\HLJLp{(}@*)(*@\HLJLn{lotka{\_}volterra}@*)(*@\HLJLp{,}@*)(*@\HLJLn{u0}@*)(*@\HLJLp{,}@*)(*@\HLJLn{tspan}@*)(*@\HLJLp{,}@*)(*@\HLJLn{p}@*)(*@\HLJLp{)}@*)
\end{lstlisting}
Then to solve the differential equations, you can simply call \texttt{solve} on the \texttt{prob}:
\begin{lstlisting}
(*@\HLJLn{sol}@*) (*@\HLJLoB{=}@*) (*@\HLJLnf{solve}@*)(*@\HLJLp{(}@*)(*@\HLJLn{prob}@*)(*@\HLJLp{)}@*)
(*@\HLJLk{using}@*) (*@\HLJLn{Plots}@*)
(*@\HLJLnf{plot}@*)(*@\HLJLp{(}@*)(*@\HLJLn{sol}@*)(*@\HLJLp{)}@*)
\end{lstlisting}

\includegraphics[width=\linewidth]{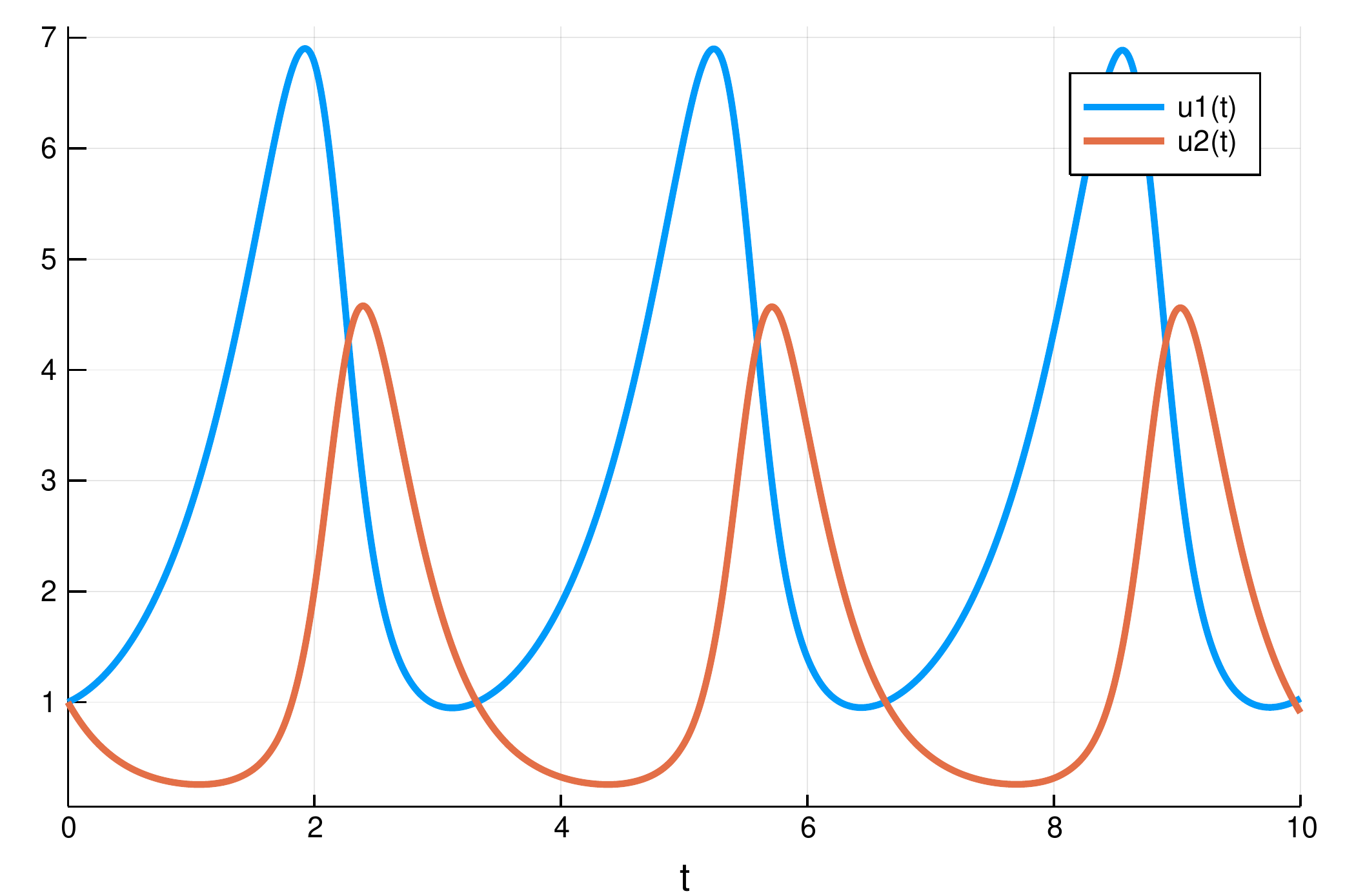}

One last thing to note is that we can make our initial condition (\texttt{u0}) and time spans (\texttt{tspans}) to be functions of the parameters (the elements of \texttt{p}). For example, we can define the \texttt{ODEProblem}:
\begin{lstlisting}
(*@\HLJLnf{u0{\_}f}@*)(*@\HLJLp{(}@*)(*@\HLJLn{p}@*)(*@\HLJLp{,}@*)(*@\HLJLn{t0}@*)(*@\HLJLp{)}@*) (*@\HLJLoB{=}@*) (*@\HLJLp{[}@*)(*@\HLJLn{p}@*)(*@\HLJLp{[}@*)(*@\HLJLni{2}@*)(*@\HLJLp{],}@*)(*@\HLJLn{p}@*)(*@\HLJLp{[}@*)(*@\HLJLni{4}@*)(*@\HLJLp{]]}@*)
(*@\HLJLnf{tspan{\_}f}@*)(*@\HLJLp{(}@*)(*@\HLJLn{p}@*)(*@\HLJLp{)}@*) (*@\HLJLoB{=}@*) (*@\HLJLp{(}@*)(*@\HLJLnfB{0.0}@*)(*@\HLJLp{,}@*)(*@\HLJLni{10}@*)(*@\HLJLoB{*}@*)(*@\HLJLn{p}@*)(*@\HLJLp{[}@*)(*@\HLJLni{4}@*)(*@\HLJLp{])}@*)
(*@\HLJLn{p}@*) (*@\HLJLoB{=}@*) (*@\HLJLp{[}@*)(*@\HLJLnfB{1.5}@*)(*@\HLJLp{,}@*)(*@\HLJLnfB{1.0}@*)(*@\HLJLp{,}@*)(*@\HLJLnfB{3.0}@*)(*@\HLJLp{,}@*)(*@\HLJLnfB{1.0}@*)(*@\HLJLp{]}@*)
(*@\HLJLn{prob}@*) (*@\HLJLoB{=}@*) (*@\HLJLnf{ODEProblem}@*)(*@\HLJLp{(}@*)(*@\HLJLn{lotka{\_}volterra}@*)(*@\HLJLp{,}@*)(*@\HLJLn{u0{\_}f}@*)(*@\HLJLp{,}@*)(*@\HLJLn{tspan{\_}f}@*)(*@\HLJLp{,}@*)(*@\HLJLn{p}@*)(*@\HLJLp{)}@*)
\end{lstlisting}

In this form, everything about the problem is determined by the parameter vector (\texttt{p}, referred to as \texttt{\ensuremath{\theta}} in associated literature). The utility of this will be seen later.

DifferentialEquations.jl has many powerful options for customising things like accuracy, tolerances, solver methods, events and more; check out \href{http://docs.juliadiffeq.org/latest/}{the docs} for more details on how to use it in more advanced ways.

\section{Let's Put an ODE Into a Neural Net Framework!}
To understand embedding an ODE into a neural network, let's look at what a neural network layer actually is. A layer is really just a \emph{differentiable function} which takes in a vector of size \texttt{n} and spits out a new vector of size \texttt{m}. That's it! Layers have traditionally been simple functions like matrix multiply, but in the spirit of \href{https://julialang.org/blog/2018/12/ml-language-compiler}{differentiable programming} people are increasingly experimenting with much more complex functions, such as ray tracers and physics engines.

Turns out that differential equations solvers fit this framework, too: A solve takes in some vector \texttt{p} (which might include parameters like the initial starting point), and outputs some new vector, the solution. Moreover it's differentiable, which means we can put it straight into a larger differentiable program. This larger program can happily include neural networks, and we can keep using standard optimisation techniques like ADAM to optimise their weights.

DiffEqFlux.jl makes it convenient to do just this; let's take it for a spin. We'll start by solving an equation as before, without gradients.
\begin{lstlisting}
(*@\HLJLn{p}@*) (*@\HLJLoB{=}@*) (*@\HLJLp{[}@*)(*@\HLJLnfB{1.5}@*)(*@\HLJLp{,}@*)(*@\HLJLnfB{1.0}@*)(*@\HLJLp{,}@*)(*@\HLJLnfB{3.0}@*)(*@\HLJLp{,}@*)(*@\HLJLnfB{1.0}@*)(*@\HLJLp{]}@*)
(*@\HLJLn{prob}@*) (*@\HLJLoB{=}@*) (*@\HLJLnf{ODEProblem}@*)(*@\HLJLp{(}@*)(*@\HLJLn{lotka{\_}volterra}@*)(*@\HLJLp{,}@*)(*@\HLJLn{u0}@*)(*@\HLJLp{,}@*)(*@\HLJLn{tspan}@*)(*@\HLJLp{,}@*)(*@\HLJLn{p}@*)(*@\HLJLp{)}@*)
(*@\HLJLn{sol}@*) (*@\HLJLoB{=}@*) (*@\HLJLnf{solve}@*)(*@\HLJLp{(}@*)(*@\HLJLn{prob}@*)(*@\HLJLp{,}@*)(*@\HLJLnf{Tsit5}@*)(*@\HLJLp{(),}@*)(*@\HLJLn{saveat}@*)(*@\HLJLoB{=}@*)(*@\HLJLnfB{0.1}@*)(*@\HLJLp{)}@*)
(*@\HLJLn{A}@*) (*@\HLJLoB{=}@*) (*@\HLJLn{sol}@*)(*@\HLJLp{[}@*)(*@\HLJLni{1}@*)(*@\HLJLp{,}@*)(*@\HLJLoB{:}@*)(*@\HLJLp{]}@*) (*@\HLJLcs{{\#} length 101 vector}@*)
\end{lstlisting}

Let's plot \texttt{(t,A)} over the ODE's solution to see what we got:
\begin{lstlisting}
(*@\HLJLnf{plot}@*)(*@\HLJLp{(}@*)(*@\HLJLn{sol}@*)(*@\HLJLp{)}@*)
(*@\HLJLn{t}@*) (*@\HLJLoB{=}@*) (*@\HLJLni{0}@*)(*@\HLJLoB{:}@*)(*@\HLJLnfB{0.1}@*)(*@\HLJLoB{:}@*)(*@\HLJLnfB{10.0}@*)
(*@\HLJLnf{scatter!}@*)(*@\HLJLp{(}@*)(*@\HLJLn{t}@*)(*@\HLJLp{,}@*)(*@\HLJLn{A}@*)(*@\HLJLp{)}@*)
\end{lstlisting}

\includegraphics[width=\linewidth]{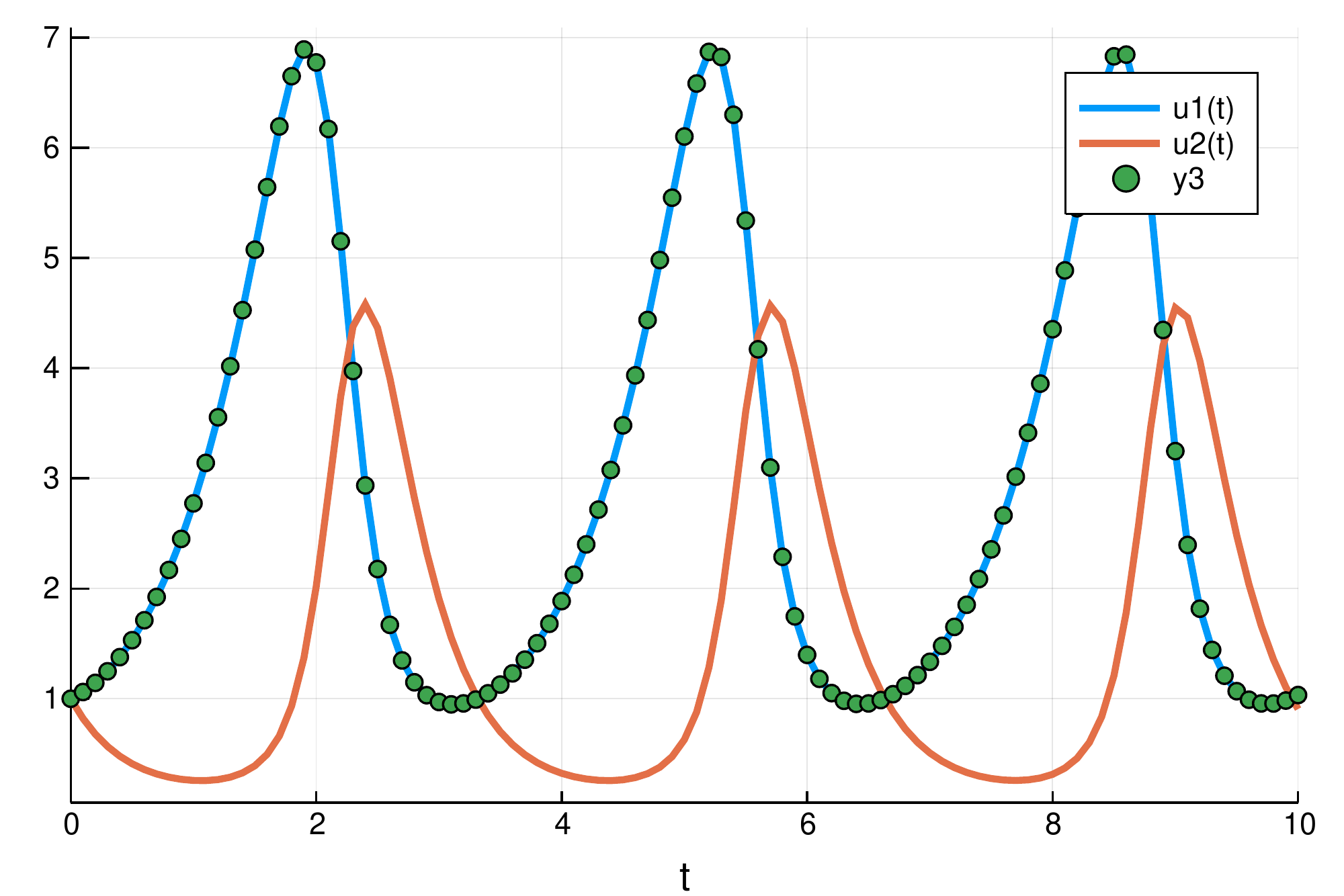}

The most basic differential equation layer is \texttt{diffeq\_rd}, which does the same thing with a slightly altered syntax. \texttt{diffeq\_rd} takes in parameters \texttt{p} for the integrand, puts it in the differential equation defined by \texttt{prob}, and solves it with the chosen arguments (solver, tolerance, etc). For example:
\begin{lstlisting}
(*@\HLJLk{using}@*) (*@\HLJLn{Flux}@*)(*@\HLJLp{,}@*) (*@\HLJLn{DiffEqFlux}@*)
(*@\HLJLnf{diffeq{\_}rd}@*)(*@\HLJLp{(}@*)(*@\HLJLn{p}@*)(*@\HLJLp{,}@*)(*@\HLJLn{prob}@*)(*@\HLJLp{,}@*)(*@\HLJLnf{Tsit5}@*)(*@\HLJLp{(),}@*)(*@\HLJLn{saveat}@*)(*@\HLJLoB{=}@*)(*@\HLJLnfB{0.1}@*)(*@\HLJLp{)}@*)
\end{lstlisting}
\begin{lstlisting}
retcode: Success
Interpolation: 1st order linear
t: 101-element Array{Float64,1}:
  0.0
  0.1
  0.2
  0.3
  0.4
  0.5
  0.6
  0.7
  0.8
  0.9
  (*@\ensuremath{\vdots}@*)  
  9.2
  9.3
  9.4
  9.5
  9.6
  9.7
  9.8
  9.9
 10.0
u: 101-element Array{Array{Float64,1},1}:
 [1.0, 1.0]         
 [1.06108, 0.821084]
 [1.14403, 0.679053]
 [1.24917, 0.566893]
 [1.37764, 0.478813]
 [1.53123, 0.410156]
 [1.71227, 0.357265]
 [1.92358, 0.317347]
 [2.16839, 0.288389]
 [2.45025, 0.269054]
 (*@\ensuremath{\vdots}@*)                  
 [1.81728, 4.06495] 
 [1.44276, 3.53974] 
 [1.20891, 2.99146] 
 [1.06859, 2.48207] 
 [0.991023, 2.03725]
 [0.957421, 1.66321]
 [0.956979, 1.35559]
 [0.983561, 1.10629]
 [1.03376, 0.906371]
\end{lstlisting}
The nice thing about \texttt{diffeq\_rd} is that it takes care of the type handling necessary to make it compatible with the neural network framework (here Flux). To show this, let's define a neural network with the function as our single layer, and then a loss function that is the squared distance of the output values from \texttt{1}. In Flux, this looks like:
\begin{lstlisting}
(*@\HLJLn{p}@*) (*@\HLJLoB{=}@*) (*@\HLJLnf{param}@*)(*@\HLJLp{([}@*)(*@\HLJLnfB{2.2}@*)(*@\HLJLp{,}@*) (*@\HLJLnfB{1.0}@*)(*@\HLJLp{,}@*) (*@\HLJLnfB{2.0}@*)(*@\HLJLp{,}@*) (*@\HLJLnfB{0.4}@*)(*@\HLJLp{])}@*) (*@\HLJLcs{{\#} Initial Parameter Vector}@*)
(*@\HLJLn{params}@*) (*@\HLJLoB{=}@*) (*@\HLJLn{Flux}@*)(*@\HLJLoB{.}@*)(*@\HLJLnf{Params}@*)(*@\HLJLp{([}@*)(*@\HLJLn{p}@*)(*@\HLJLp{])}@*)

(*@\HLJLk{function}@*) (*@\HLJLnf{predict{\_}rd}@*)(*@\HLJLp{()}@*) (*@\HLJLcs{{\#} Our 1-layer neural network}@*)
  (*@\HLJLnf{diffeq{\_}rd}@*)(*@\HLJLp{(}@*)(*@\HLJLn{p}@*)(*@\HLJLp{,}@*)(*@\HLJLn{prob}@*)(*@\HLJLp{,}@*)(*@\HLJLnf{Tsit5}@*)(*@\HLJLp{(),}@*)(*@\HLJLn{saveat}@*)(*@\HLJLoB{=}@*)(*@\HLJLnfB{0.1}@*)(*@\HLJLp{)[}@*)(*@\HLJLni{1}@*)(*@\HLJLp{,}@*)(*@\HLJLoB{:}@*)(*@\HLJLp{]}@*)
(*@\HLJLk{end}@*)

(*@\HLJLnf{loss{\_}rd}@*)(*@\HLJLp{()}@*) (*@\HLJLoB{=}@*) (*@\HLJLnf{sum}@*)(*@\HLJLp{(}@*)(*@\HLJLn{abs2}@*)(*@\HLJLp{,}@*)(*@\HLJLn{x}@*)(*@\HLJLoB{-}@*)(*@\HLJLni{1}@*) (*@\HLJLk{for}@*) (*@\HLJLn{x}@*) (*@\HLJLkp{in}@*) (*@\HLJLnf{predict{\_}rd}@*)(*@\HLJLp{())}@*) (*@\HLJLcs{{\#} loss function}@*)
\end{lstlisting}
Now we tell Flux to train the neural network by running a 100 epoch to minimise our loss function (\texttt{loss\_rd()}) and thus obtain the optimized parameters:
\begin{lstlisting}
(*@\HLJLn{data}@*) (*@\HLJLoB{=}@*) (*@\HLJLn{Iterators}@*)(*@\HLJLoB{.}@*)(*@\HLJLnf{repeated}@*)(*@\HLJLp{((),}@*) (*@\HLJLni{100}@*)(*@\HLJLp{)}@*)
(*@\HLJLn{opt}@*) (*@\HLJLoB{=}@*) (*@\HLJLnf{ADAM}@*)(*@\HLJLp{(}@*)(*@\HLJLnfB{0.1}@*)(*@\HLJLp{)}@*)
(*@\HLJLn{cb}@*) (*@\HLJLoB{=}@*) (*@\HLJLk{function}@*) (*@\HLJLp{()}@*) (*@\HLJLcs{{\#}callback function to observe training}@*)
  (*@\HLJLcs{{\#}display(loss{\_}rd())}@*)
  (*@\HLJLcs{{\#} using {\textasciigrave}remake{\textasciigrave} to re-create our {\textasciigrave}prob{\textasciigrave} with current parameters {\textasciigrave}p{\textasciigrave}}@*)
  (*@\HLJLcs{{\#} Creates an animation}@*)
  (*@\HLJLcs{{\#}display(plot(solve(remake(prob,p=Flux.data(p)),Tsit5(),saveat=0.1),ylim=(0,6)))}@*)
(*@\HLJLk{end}@*)

(*@\HLJLcs{{\#} Display the ODE with the initial parameter values.}@*)
(*@\HLJLnf{cb}@*)(*@\HLJLp{()}@*)

(*@\HLJLn{Flux}@*)(*@\HLJLoB{.}@*)(*@\HLJLnf{train!}@*)(*@\HLJLp{(}@*)(*@\HLJLn{loss{\_}rd}@*)(*@\HLJLp{,}@*) (*@\HLJLn{params}@*)(*@\HLJLp{,}@*) (*@\HLJLn{data}@*)(*@\HLJLp{,}@*) (*@\HLJLn{opt}@*)(*@\HLJLp{,}@*) (*@\HLJLn{cb}@*) (*@\HLJLoB{=}@*) (*@\HLJLn{cb}@*)(*@\HLJLp{)}@*)

(*@\HLJLnf{plot}@*)(*@\HLJLp{(}@*)(*@\HLJLnf{solve}@*)(*@\HLJLp{(}@*)(*@\HLJLnf{remake}@*)(*@\HLJLp{(}@*)(*@\HLJLn{prob}@*)(*@\HLJLp{,}@*)(*@\HLJLn{p}@*)(*@\HLJLoB{=}@*)(*@\HLJLn{Flux}@*)(*@\HLJLoB{.}@*)(*@\HLJLnf{data}@*)(*@\HLJLp{(}@*)(*@\HLJLn{p}@*)(*@\HLJLp{)),}@*)(*@\HLJLnf{Tsit5}@*)(*@\HLJLp{(),}@*)(*@\HLJLn{saveat}@*)(*@\HLJLoB{=}@*)(*@\HLJLnfB{0.1}@*)(*@\HLJLp{),}@*)(*@\HLJLn{ylim}@*)(*@\HLJLoB{=}@*)(*@\HLJLp{(}@*)(*@\HLJLni{0}@*)(*@\HLJLp{,}@*)(*@\HLJLni{6}@*)(*@\HLJLp{))}@*)
\end{lstlisting}

\includegraphics[width=\linewidth]{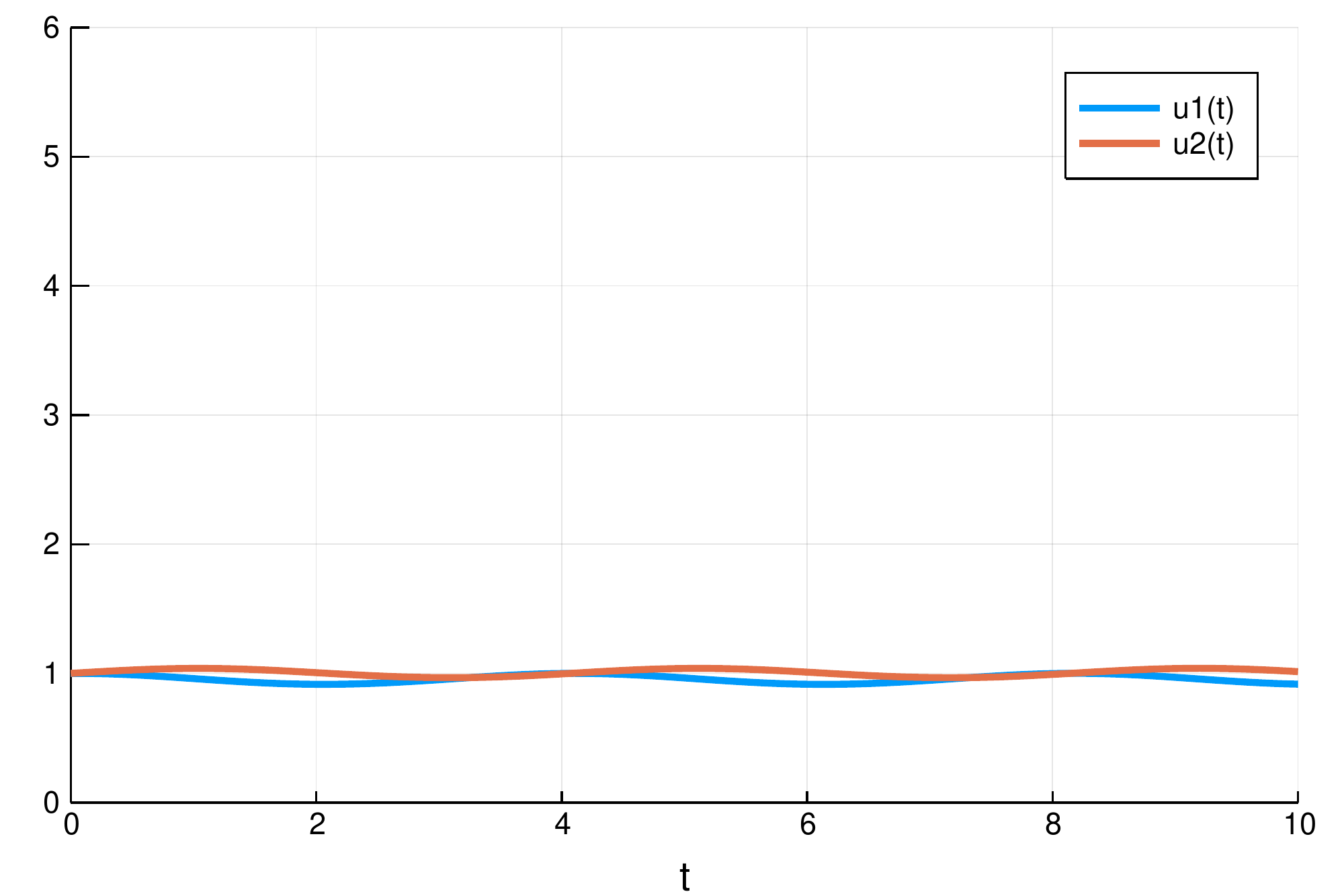}

(Animation omitted from the paper. Please see the \href{https://julialang.org/blog/2019/01/fluxdiffeq}{original blog post})

\href{https://github.com/FluxML/model-zoo/blob/da4156b4a9fb0d5907dcb6e21d0e78c72b6122e0/other/diffeq/ode.jl}{This code can be found in the model-zoo}

Flux finds the parameters of the neural network (\texttt{p}) which minimize the cost function, i.e. it trains the neural network: it just so happens that the forward pass of the neural network includes solving an ODE. Since our cost function put a penalty whenever the number of rabbits was far from 1, our neural network found parameters where our population of rabbits and wolves are both constant 1.

Now that we have solving ODEs as just a layer, we can add it anywhere. For example, the multilayer perceptron is written in Flux as

\begin{lstlisting}
(*@\HLJLn{m}@*) (*@\HLJLoB{=}@*) (*@\HLJLnf{Chain}@*)(*@\HLJLp{(}@*)
  (*@\HLJLnf{Dense}@*)(*@\HLJLp{(}@*)(*@\HLJLni{28}@*)(*@\HLJLoB{{\textasciicircum}}@*)(*@\HLJLni{2}@*)(*@\HLJLp{,}@*) (*@\HLJLni{32}@*)(*@\HLJLp{,}@*) (*@\HLJLn{relu}@*)(*@\HLJLp{),}@*)
  (*@\HLJLnf{Dense}@*)(*@\HLJLp{(}@*)(*@\HLJLni{32}@*)(*@\HLJLp{,}@*) (*@\HLJLni{10}@*)(*@\HLJLp{),}@*)
  (*@\HLJLn{softmax}@*)(*@\HLJLp{)}@*)
\end{lstlisting}

and if we had an appropriate ODE which took a parameter vector of the right size, we can stick it right in there:

\begin{lstlisting}
(*@\HLJLn{m}@*) (*@\HLJLoB{=}@*) (*@\HLJLnf{Chain}@*)(*@\HLJLp{(}@*)
  (*@\HLJLnf{Dense}@*)(*@\HLJLp{(}@*)(*@\HLJLni{28}@*)(*@\HLJLoB{{\textasciicircum}}@*)(*@\HLJLni{2}@*)(*@\HLJLp{,}@*) (*@\HLJLni{32}@*)(*@\HLJLp{,}@*) (*@\HLJLn{relu}@*)(*@\HLJLp{),}@*)
  (*@\HLJLcs{{\#} this would require an ODE of 32 parameters}@*)
  (*@\HLJLn{p}@*) (*@\HLJLoB{->}@*) (*@\HLJLnf{diffeq{\_}rd}@*)(*@\HLJLp{(}@*)(*@\HLJLn{p}@*)(*@\HLJLp{,}@*)(*@\HLJLn{prob}@*)(*@\HLJLp{,}@*)(*@\HLJLnf{Tsit5}@*)(*@\HLJLp{(),}@*)(*@\HLJLn{saveat}@*)(*@\HLJLoB{=}@*)(*@\HLJLnfB{0.1}@*)(*@\HLJLp{)[}@*)(*@\HLJLni{1}@*)(*@\HLJLp{,}@*)(*@\HLJLoB{:}@*)(*@\HLJLp{],}@*)
  (*@\HLJLnf{Dense}@*)(*@\HLJLp{(}@*)(*@\HLJLni{32}@*)(*@\HLJLp{,}@*) (*@\HLJLni{10}@*)(*@\HLJLp{),}@*)
  (*@\HLJLn{softmax}@*)(*@\HLJLp{)}@*)
\end{lstlisting}

or we can stick it into a convolutional neural network, where the previous layers define the initial condition for the ODE:

\begin{lstlisting}
(*@\HLJLn{m}@*) (*@\HLJLoB{=}@*) (*@\HLJLnf{Chain}@*)(*@\HLJLp{(}@*)
  (*@\HLJLnf{Conv}@*)(*@\HLJLp{((}@*)(*@\HLJLni{2}@*)(*@\HLJLp{,}@*)(*@\HLJLni{2}@*)(*@\HLJLp{),}@*) (*@\HLJLni{1}@*)(*@\HLJLoB{=>}@*)(*@\HLJLni{16}@*)(*@\HLJLp{,}@*) (*@\HLJLn{relu}@*)(*@\HLJLp{),}@*)
  (*@\HLJLn{x}@*) (*@\HLJLoB{->}@*) (*@\HLJLnf{maxpool}@*)(*@\HLJLp{(}@*)(*@\HLJLn{x}@*)(*@\HLJLp{,}@*) (*@\HLJLp{(}@*)(*@\HLJLni{2}@*)(*@\HLJLp{,}@*)(*@\HLJLni{2}@*)(*@\HLJLp{)),}@*)
  (*@\HLJLnf{Conv}@*)(*@\HLJLp{((}@*)(*@\HLJLni{2}@*)(*@\HLJLp{,}@*)(*@\HLJLni{2}@*)(*@\HLJLp{),}@*) (*@\HLJLni{16}@*)(*@\HLJLoB{=>}@*)(*@\HLJLni{8}@*)(*@\HLJLp{,}@*) (*@\HLJLn{relu}@*)(*@\HLJLp{),}@*)
  (*@\HLJLn{x}@*) (*@\HLJLoB{->}@*) (*@\HLJLnf{maxpool}@*)(*@\HLJLp{(}@*)(*@\HLJLn{x}@*)(*@\HLJLp{,}@*) (*@\HLJLp{(}@*)(*@\HLJLni{2}@*)(*@\HLJLp{,}@*)(*@\HLJLni{2}@*)(*@\HLJLp{)),}@*)
  (*@\HLJLn{x}@*) (*@\HLJLoB{->}@*) (*@\HLJLnf{reshape}@*)(*@\HLJLp{(}@*)(*@\HLJLn{x}@*)(*@\HLJLp{,}@*) (*@\HLJLoB{:}@*)(*@\HLJLp{,}@*) (*@\HLJLnf{size}@*)(*@\HLJLp{(}@*)(*@\HLJLn{x}@*)(*@\HLJLp{,}@*) (*@\HLJLni{4}@*)(*@\HLJLp{)),}@*)
  (*@\HLJLn{x}@*) (*@\HLJLoB{->}@*) (*@\HLJLnf{diffeq{\_}rd}@*)(*@\HLJLp{(}@*)(*@\HLJLn{p}@*)(*@\HLJLp{,}@*)(*@\HLJLn{prob}@*)(*@\HLJLp{,}@*)(*@\HLJLnf{Tsit5}@*)(*@\HLJLp{(),}@*)(*@\HLJLn{saveat}@*)(*@\HLJLoB{=}@*)(*@\HLJLnfB{0.1}@*)(*@\HLJLp{,}@*)(*@\HLJLn{u0}@*)(*@\HLJLoB{=}@*)(*@\HLJLn{x}@*)(*@\HLJLp{)[}@*)(*@\HLJLni{1}@*)(*@\HLJLp{,}@*)(*@\HLJLoB{:}@*)(*@\HLJLp{],}@*)
  (*@\HLJLnf{Dense}@*)(*@\HLJLp{(}@*)(*@\HLJLni{288}@*)(*@\HLJLp{,}@*) (*@\HLJLni{10}@*)(*@\HLJLp{),}@*) (*@\HLJLn{softmax}@*)(*@\HLJLp{)}@*) (*@\HLJLoB{|>}@*) (*@\HLJLn{gpu}@*)
\end{lstlisting}

As long as you can write down the forward pass, we can take any parameterised, differentiable program and optimise it. The world is your oyster.

\section{Why is a full ODE solver suite necessary for doing this well?}
Where we have combined an existing solver suite and deep learning library, the excellent \href{https://github.com/rtqichen/torchdiffeq}{torchdiffeq} project takes an alternative approach, instead implementing solver methods directly in PyTorch, including an adaptive Runge Kutta 4-5 (\texttt{dopri5}) and an Adams-Bashforth-Moulton method (\texttt{adams}). However, while their approach is very effective for certain kinds of models, not having access to a full solver suite is limiting.

Consider the following example, the \href{https://www.radford.edu/~thompson/vodef90web/problems/demosnodislin/Single/DemoRobertson/demorobertson.pdf}{ROBER ODE}. The most well-tested (and optimized) implementation of an Adams-Bashforth-Moulton method is the \href{https://computation.llnl.gov/projects/sundials}{CVODE integrator in the C++ package SUNDIALS} (a derivative of the classic LSODE). Let's use DifferentialEquations.jl to call CVODE with its Adams method and have it solve the ODE for us:

\begin{lstlisting}
(*@\HLJLk{using}@*) (*@\HLJLn{ParameterizedFunctions}@*)
(*@\HLJLn{rober}@*) (*@\HLJLoB{=}@*) (*@\HLJLnd{@ode{\_}def}@*) (*@\HLJLn{Rober}@*) (*@\HLJLk{begin}@*)
  (*@\HLJLn{dy\ensuremath{\_1}}@*) (*@\HLJLoB{=}@*) (*@\HLJLoB{-}@*)(*@\HLJLn{k\ensuremath{\_1}}@*)(*@\HLJLoB{*}@*)(*@\HLJLn{y\ensuremath{\_1}}@*)(*@\HLJLoB{+}@*)(*@\HLJLn{k\ensuremath{\_3}}@*)(*@\HLJLoB{*}@*)(*@\HLJLn{y\ensuremath{\_2}}@*)(*@\HLJLoB{*}@*)(*@\HLJLn{y\ensuremath{\_3}}@*)
  (*@\HLJLn{dy\ensuremath{\_2}}@*) (*@\HLJLoB{=}@*)  (*@\HLJLn{k\ensuremath{\_1}}@*)(*@\HLJLoB{*}@*)(*@\HLJLn{y\ensuremath{\_1}}@*)(*@\HLJLoB{-}@*)(*@\HLJLn{k\ensuremath{\_2}}@*)(*@\HLJLoB{*}@*)(*@\HLJLn{y\ensuremath{\_2}}@*)(*@\HLJLoB{{\textasciicircum}}@*)(*@\HLJLni{2}@*)(*@\HLJLoB{-}@*)(*@\HLJLn{k\ensuremath{\_3}}@*)(*@\HLJLoB{*}@*)(*@\HLJLn{y\ensuremath{\_2}}@*)(*@\HLJLoB{*}@*)(*@\HLJLn{y\ensuremath{\_3}}@*)
  (*@\HLJLn{dy\ensuremath{\_3}}@*) (*@\HLJLoB{=}@*)  (*@\HLJLn{k\ensuremath{\_2}}@*)(*@\HLJLoB{*}@*)(*@\HLJLn{y\ensuremath{\_2}}@*)(*@\HLJLoB{{\textasciicircum}}@*)(*@\HLJLni{2}@*)
(*@\HLJLk{end}@*) (*@\HLJLn{k\ensuremath{\_1}}@*) (*@\HLJLn{k\ensuremath{\_2}}@*) (*@\HLJLn{k\ensuremath{\_3}}@*)
(*@\HLJLn{prob}@*) (*@\HLJLoB{=}@*) (*@\HLJLnf{ODEProblem}@*)(*@\HLJLp{(}@*)(*@\HLJLn{rober}@*)(*@\HLJLp{,[}@*)(*@\HLJLnfB{1.0}@*)(*@\HLJLp{;}@*)(*@\HLJLnfB{0.0}@*)(*@\HLJLp{;}@*)(*@\HLJLnfB{0.0}@*)(*@\HLJLp{],(}@*)(*@\HLJLnfB{0.0}@*)(*@\HLJLp{,}@*)(*@\HLJLnfB{1e11}@*)(*@\HLJLp{),(}@*)(*@\HLJLnfB{0.04}@*)(*@\HLJLp{,}@*)(*@\HLJLnfB{3e7}@*)(*@\HLJLp{,}@*)(*@\HLJLnfB{1e4}@*)(*@\HLJLp{))}@*)
(*@\HLJLcs{{\#} solve(prob,CVODE{\_}Adams()) {\#} Takes forever!}@*)
\end{lstlisting}

(For those familiar with solving ODEs in MATLAB, this is similar to \texttt{ode113})

Both this and the \texttt{dopri} method from \href{https://www.unige.ch/~hairer/software.html}{Ernst Hairer's Fortran Suite} stall and fail to solve the equation. This happens because the ODE is \href{https://en.wikipedia.org/wiki/Stiff_equation}{stiff}, and thus methods with "smaller stability regions" will not be able to solve it appropriately (for more details, I suggest reading Hairer's Solving Ordinary Differential Equations II). On the other hand \texttt{Rodas5()} to this problem, the equation is solved in a blink of an eye:

\begin{lstlisting}
(*@\HLJLn{sol}@*) (*@\HLJLoB{=}@*) (*@\HLJLnf{solve}@*)(*@\HLJLp{(}@*)(*@\HLJLn{prob}@*)(*@\HLJLp{,}@*)(*@\HLJLnf{Rodas5}@*)(*@\HLJLp{())}@*)
(*@\HLJLk{using}@*) (*@\HLJLn{Plots}@*)
(*@\HLJLnf{plot}@*)(*@\HLJLp{(}@*)(*@\HLJLn{sol}@*)(*@\HLJLp{,}@*)(*@\HLJLn{xscale}@*)(*@\HLJLoB{=:}@*)(*@\HLJLn{log10}@*)(*@\HLJLp{,}@*)(*@\HLJLn{tspan}@*)(*@\HLJLoB{=}@*)(*@\HLJLp{(}@*)(*@\HLJLnfB{0.1}@*)(*@\HLJLp{,}@*)(*@\HLJLnfB{1e11}@*)(*@\HLJLp{))}@*)
\end{lstlisting}

\includegraphics[width=\linewidth]{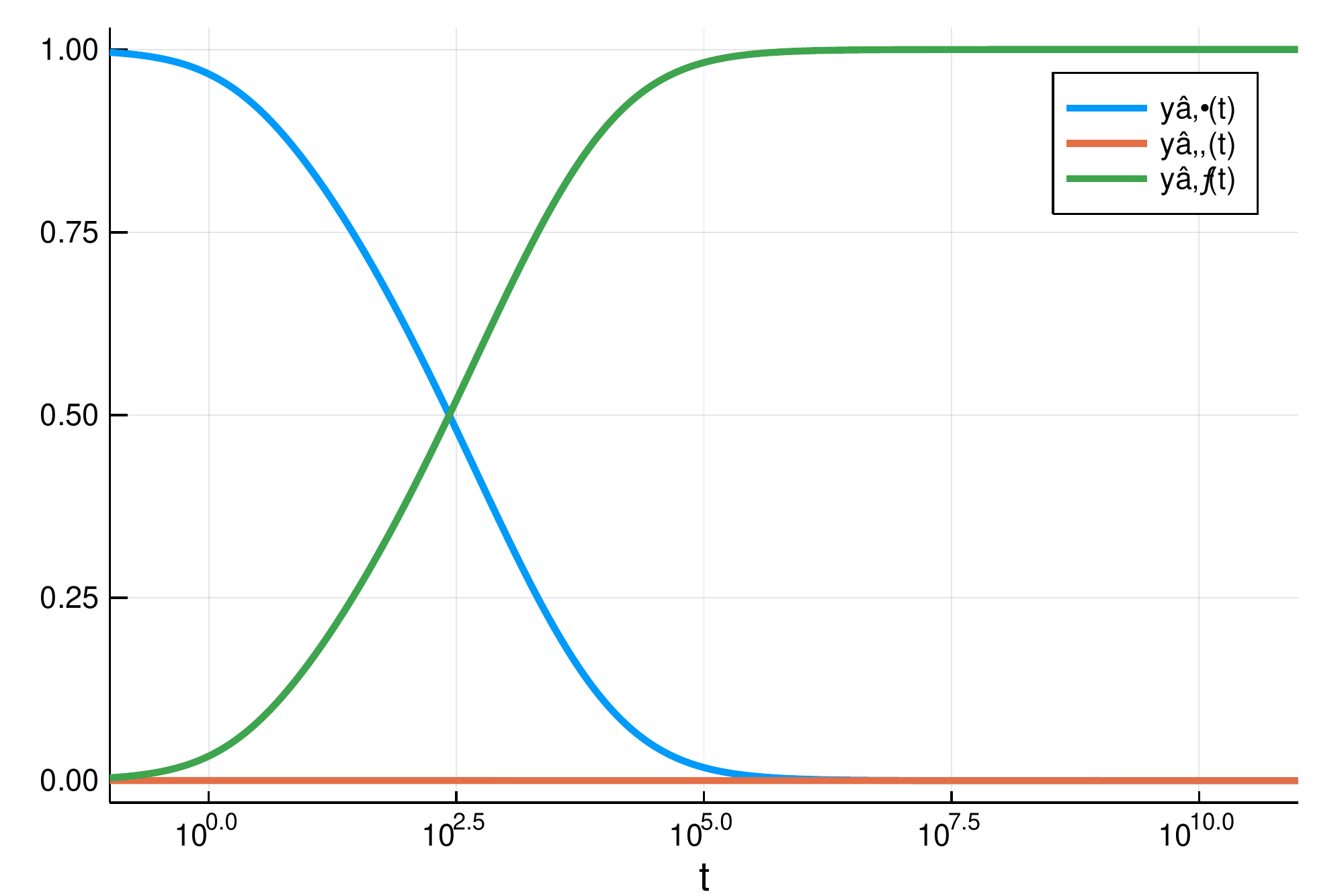}

This is just one example of subtlety in integration: Stabilizing explicit methods via PI-adaptive controllers, step prediction in implicit solvers, etc. are all intricate details that take a lot of time and testing to become efficient and robust. Different problems require different methods: \href{http://docs.juliadiffeq.org/latest/solvers/dynamical_solve.html#Symplectic-Integrators-1}{Symplectic integrators} are required to \href{https://scicomp.stackexchange.com/questions/29149/what-does-symplectic-mean-in-reference-to-numerical-integrators-and-does-scip/29154#29154}{adequately handle physical many problems without drift}, and tools like \href{http://docs.juliadiffeq.org/latest/solvers/split_ode_solve.html#Implicit-Explicit-(IMEX)-ODE-1}{IMEX integrators} are required to handle ODEs which \href{https://www.youtube.com/watch?v=okGybBmihOE}{come from partial differential equations}. Building a production-quality solver is thus an enormous undertaking and relatively few exist.

Rather than building an ML-specific solver suite in parallel to one suitable for scientific computing, in Julia they are one and the same, meaning you can take advantage of all of these methods today.

\section{What kinds of differential equations are there?}
Ordinary differential equations are only one kind of differential equation. There are many additional features you can add to the structure of a differential equation. For example, the amount of bunnies in the future isn't dependent on the number of bunnies right now because it takes a non-zero amount of time for a parent to come to term after a child is incepted. Thus the birth rate of bunnies is actually due to the amount of bunnies in the past. Using a lag term in a differential equation's derivative makes this equation known as a delay differential equation (DDE). Since \href{http://docs.juliadiffeq.org/latest/tutorials/dde_example.html}{DifferentialEquations.jl handles DDEs} through the same interface as ODEs, it can be used as a layer in Flux as well. Here's an example:

\begin{lstlisting}
(*@\HLJLk{function}@*) (*@\HLJLnf{delay{\_}lotka{\_}volterra}@*)(*@\HLJLp{(}@*)(*@\HLJLn{du}@*)(*@\HLJLp{,}@*)(*@\HLJLn{u}@*)(*@\HLJLp{,}@*)(*@\HLJLn{h}@*)(*@\HLJLp{,}@*)(*@\HLJLn{p}@*)(*@\HLJLp{,}@*)(*@\HLJLn{t}@*)(*@\HLJLp{)}@*)
  (*@\HLJLn{x}@*)(*@\HLJLp{,}@*) (*@\HLJLn{y}@*) (*@\HLJLoB{=}@*) (*@\HLJLn{u}@*)
  (*@\HLJLn{\ensuremath{\alpha}}@*)(*@\HLJLp{,}@*) (*@\HLJLn{\ensuremath{\beta}}@*)(*@\HLJLp{,}@*) (*@\HLJLn{\ensuremath{\delta}}@*)(*@\HLJLp{,}@*) (*@\HLJLn{\ensuremath{\gamma}}@*) (*@\HLJLoB{=}@*) (*@\HLJLn{p}@*)
  (*@\HLJLn{du}@*)(*@\HLJLp{[}@*)(*@\HLJLni{1}@*)(*@\HLJLp{]}@*) (*@\HLJLoB{=}@*) (*@\HLJLn{dx}@*) (*@\HLJLoB{=}@*) (*@\HLJLp{(}@*)(*@\HLJLn{\ensuremath{\alpha}}@*) (*@\HLJLoB{-}@*) (*@\HLJLn{\ensuremath{\beta}}@*)(*@\HLJLoB{*}@*)(*@\HLJLn{y}@*)(*@\HLJLp{)}@*)(*@\HLJLoB{*}@*)(*@\HLJLnf{h}@*)(*@\HLJLp{(}@*)(*@\HLJLn{p}@*)(*@\HLJLp{,}@*)(*@\HLJLn{t}@*)(*@\HLJLoB{-}@*)(*@\HLJLnfB{0.1}@*)(*@\HLJLp{)[}@*)(*@\HLJLni{1}@*)(*@\HLJLp{]}@*)
  (*@\HLJLn{du}@*)(*@\HLJLp{[}@*)(*@\HLJLni{2}@*)(*@\HLJLp{]}@*) (*@\HLJLoB{=}@*) (*@\HLJLn{dy}@*) (*@\HLJLoB{=}@*) (*@\HLJLp{(}@*)(*@\HLJLn{\ensuremath{\delta}}@*)(*@\HLJLoB{*}@*)(*@\HLJLn{x}@*) (*@\HLJLoB{-}@*) (*@\HLJLn{\ensuremath{\gamma}}@*)(*@\HLJLp{)}@*)(*@\HLJLoB{*}@*)(*@\HLJLn{y}@*)
(*@\HLJLk{end}@*)
(*@\HLJLnf{h}@*)(*@\HLJLp{(}@*)(*@\HLJLn{p}@*)(*@\HLJLp{,}@*)(*@\HLJLn{t}@*)(*@\HLJLp{)}@*) (*@\HLJLoB{=}@*) (*@\HLJLnf{ones}@*)(*@\HLJLp{(}@*)(*@\HLJLnf{eltype}@*)(*@\HLJLp{(}@*)(*@\HLJLn{p}@*)(*@\HLJLp{),}@*)(*@\HLJLni{2}@*)(*@\HLJLp{)}@*)
(*@\HLJLn{prob}@*) (*@\HLJLoB{=}@*) (*@\HLJLnf{DDEProblem}@*)(*@\HLJLp{(}@*)(*@\HLJLn{delay{\_}lotka{\_}volterra}@*)(*@\HLJLp{,[}@*)(*@\HLJLnfB{1.0}@*)(*@\HLJLp{,}@*)(*@\HLJLnfB{1.0}@*)(*@\HLJLp{],}@*)(*@\HLJLn{h}@*)(*@\HLJLp{,(}@*)(*@\HLJLnfB{0.0}@*)(*@\HLJLp{,}@*)(*@\HLJLnfB{10.0}@*)(*@\HLJLp{),}@*)(*@\HLJLn{constant{\_}lags}@*)(*@\HLJLoB{=}@*)(*@\HLJLp{[}@*)(*@\HLJLnfB{0.1}@*)(*@\HLJLp{])}@*)

(*@\HLJLn{p}@*) (*@\HLJLoB{=}@*) (*@\HLJLnf{param}@*)(*@\HLJLp{([}@*)(*@\HLJLnfB{2.2}@*)(*@\HLJLp{,}@*) (*@\HLJLnfB{1.0}@*)(*@\HLJLp{,}@*) (*@\HLJLnfB{2.0}@*)(*@\HLJLp{,}@*) (*@\HLJLnfB{0.4}@*)(*@\HLJLp{])}@*)
(*@\HLJLn{params}@*) (*@\HLJLoB{=}@*) (*@\HLJLn{Flux}@*)(*@\HLJLoB{.}@*)(*@\HLJLnf{Params}@*)(*@\HLJLp{([}@*)(*@\HLJLn{p}@*)(*@\HLJLp{])}@*)
(*@\HLJLk{function}@*) (*@\HLJLnf{predict{\_}rd{\_}dde}@*)(*@\HLJLp{()}@*)
  (*@\HLJLnf{diffeq{\_}rd}@*)(*@\HLJLp{(}@*)(*@\HLJLn{p}@*)(*@\HLJLp{,}@*)(*@\HLJLn{prob}@*)(*@\HLJLp{,}@*)(*@\HLJLnf{MethodOfSteps}@*)(*@\HLJLp{(}@*)(*@\HLJLnf{Tsit5}@*)(*@\HLJLp{()),}@*)(*@\HLJLn{saveat}@*)(*@\HLJLoB{=}@*)(*@\HLJLnfB{0.1}@*)(*@\HLJLp{)[}@*)(*@\HLJLni{1}@*)(*@\HLJLp{,}@*)(*@\HLJLoB{:}@*)(*@\HLJLp{]}@*)
(*@\HLJLk{end}@*)
(*@\HLJLnf{loss{\_}rd{\_}dde}@*)(*@\HLJLp{()}@*) (*@\HLJLoB{=}@*) (*@\HLJLnf{sum}@*)(*@\HLJLp{(}@*)(*@\HLJLn{abs2}@*)(*@\HLJLp{,}@*)(*@\HLJLn{x}@*)(*@\HLJLoB{-}@*)(*@\HLJLni{1}@*) (*@\HLJLk{for}@*) (*@\HLJLn{x}@*) (*@\HLJLkp{in}@*) (*@\HLJLnf{predict{\_}rd{\_}dde}@*)(*@\HLJLp{())}@*)
(*@\HLJLnf{loss{\_}rd{\_}dde}@*)(*@\HLJLp{()}@*)
\end{lstlisting}

\begin{lstlisting}
72.94371657453573 (tracked)
\end{lstlisting}

The full code for this example, including generating an animation, \href{https://github.com/FluxML/model-zoo/blob/da4156b4a9fb0d5907dcb6e21d0e78c72b6122e0/other/diffeq/dde.jl}{can be found in the model-zoo}

Additionally we can add randomness to our differential equation to simulate how random events can cause extra births or more deaths than expected. This kind of equation is known as a stochastic differential equation (SDE). Since \href{http://docs.juliadiffeq.org/latest/tutorials/sde_example.html}{DifferentialEquations.jl handles SDEs} (and is currently the only library with adaptive stiff and non-stiff SDE integrators), these can be handled as a layer in Flux similarly. Here's a neural net layer with an SDE:

\begin{lstlisting}
(*@\HLJLk{function}@*) (*@\HLJLnf{lotka{\_}volterra{\_}noise}@*)(*@\HLJLp{(}@*)(*@\HLJLn{du}@*)(*@\HLJLp{,}@*)(*@\HLJLn{u}@*)(*@\HLJLp{,}@*)(*@\HLJLn{p}@*)(*@\HLJLp{,}@*)(*@\HLJLn{t}@*)(*@\HLJLp{)}@*)
  (*@\HLJLn{du}@*)(*@\HLJLp{[}@*)(*@\HLJLni{1}@*)(*@\HLJLp{]}@*) (*@\HLJLoB{=}@*) (*@\HLJLnfB{0.1}@*)(*@\HLJLn{u}@*)(*@\HLJLp{[}@*)(*@\HLJLni{1}@*)(*@\HLJLp{]}@*)
  (*@\HLJLn{du}@*)(*@\HLJLp{[}@*)(*@\HLJLni{2}@*)(*@\HLJLp{]}@*) (*@\HLJLoB{=}@*) (*@\HLJLnfB{0.1}@*)(*@\HLJLn{u}@*)(*@\HLJLp{[}@*)(*@\HLJLni{2}@*)(*@\HLJLp{]}@*)
(*@\HLJLk{end}@*)
(*@\HLJLn{prob}@*) (*@\HLJLoB{=}@*) (*@\HLJLnf{SDEProblem}@*)(*@\HLJLp{(}@*)(*@\HLJLn{lotka{\_}volterra}@*)(*@\HLJLp{,}@*)(*@\HLJLn{lotka{\_}volterra{\_}noise}@*)(*@\HLJLp{,[}@*)(*@\HLJLnfB{1.0}@*)(*@\HLJLp{,}@*)(*@\HLJLnfB{1.0}@*)(*@\HLJLp{],(}@*)(*@\HLJLnfB{0.0}@*)(*@\HLJLp{,}@*)(*@\HLJLnfB{10.0}@*)(*@\HLJLp{))}@*)

(*@\HLJLn{p}@*) (*@\HLJLoB{=}@*) (*@\HLJLnf{param}@*)(*@\HLJLp{([}@*)(*@\HLJLnfB{2.2}@*)(*@\HLJLp{,}@*) (*@\HLJLnfB{1.0}@*)(*@\HLJLp{,}@*) (*@\HLJLnfB{2.0}@*)(*@\HLJLp{,}@*) (*@\HLJLnfB{0.4}@*)(*@\HLJLp{])}@*)
(*@\HLJLn{params}@*) (*@\HLJLoB{=}@*) (*@\HLJLn{Flux}@*)(*@\HLJLoB{.}@*)(*@\HLJLnf{Params}@*)(*@\HLJLp{([}@*)(*@\HLJLn{p}@*)(*@\HLJLp{])}@*)
(*@\HLJLk{function}@*) (*@\HLJLnf{predict{\_}fd{\_}sde}@*)(*@\HLJLp{()}@*)
  (*@\HLJLnf{diffeq{\_}fd}@*)(*@\HLJLp{(}@*)(*@\HLJLn{p}@*)(*@\HLJLp{,}@*)(*@\HLJLn{sol}@*)(*@\HLJLoB{->}@*)(*@\HLJLn{sol}@*)(*@\HLJLp{[}@*)(*@\HLJLni{1}@*)(*@\HLJLp{,}@*)(*@\HLJLoB{:}@*)(*@\HLJLp{],}@*)(*@\HLJLni{101}@*)(*@\HLJLp{,}@*)(*@\HLJLn{prob}@*)(*@\HLJLp{,}@*)(*@\HLJLnf{SOSRI}@*)(*@\HLJLp{(),}@*)(*@\HLJLn{saveat}@*)(*@\HLJLoB{=}@*)(*@\HLJLnfB{0.1}@*)(*@\HLJLp{)}@*)
(*@\HLJLk{end}@*)
(*@\HLJLnf{loss{\_}fd{\_}sde}@*)(*@\HLJLp{()}@*) (*@\HLJLoB{=}@*) (*@\HLJLnf{sum}@*)(*@\HLJLp{(}@*)(*@\HLJLn{abs2}@*)(*@\HLJLp{,}@*)(*@\HLJLn{x}@*)(*@\HLJLoB{-}@*)(*@\HLJLni{1}@*) (*@\HLJLk{for}@*) (*@\HLJLn{x}@*) (*@\HLJLkp{in}@*) (*@\HLJLnf{predict{\_}fd{\_}sde}@*)(*@\HLJLp{())}@*)
(*@\HLJLnf{loss{\_}fd{\_}sde}@*)(*@\HLJLp{()}@*)
\end{lstlisting}

\begin{lstlisting}
4507.619695527805 (tracked)
\end{lstlisting}

And we can train the neural net to watch it in action and find parameters to make the amount of bunnies close to constant:

\begin{lstlisting}
(*@\HLJLn{data}@*) (*@\HLJLoB{=}@*) (*@\HLJLn{Iterators}@*)(*@\HLJLoB{.}@*)(*@\HLJLnf{repeated}@*)(*@\HLJLp{((),}@*) (*@\HLJLni{100}@*)(*@\HLJLp{)}@*)
(*@\HLJLn{opt}@*) (*@\HLJLoB{=}@*) (*@\HLJLnf{ADAM}@*)(*@\HLJLp{(}@*)(*@\HLJLnfB{0.1}@*)(*@\HLJLp{)}@*)
(*@\HLJLn{cb}@*) (*@\HLJLoB{=}@*) (*@\HLJLk{function}@*) (*@\HLJLp{()}@*)
  (*@\HLJLcs{{\#} display(loss{\_}fd{\_}sde())}@*)
  (*@\HLJLcs{{\#} Creates an animation}@*)
  (*@\HLJLcs{{\#}display(plot(solve(remake(prob,p=Flux.data(p)),SOSRI(),saveat=0.1),ylim=(0,6)))}@*)
(*@\HLJLk{end}@*)

(*@\HLJLcs{{\#} Display the ODE with the current parameter values.}@*)
(*@\HLJLnf{cb}@*)(*@\HLJLp{()}@*)

(*@\HLJLn{Flux}@*)(*@\HLJLoB{.}@*)(*@\HLJLnf{train!}@*)(*@\HLJLp{(}@*)(*@\HLJLn{loss{\_}fd{\_}sde}@*)(*@\HLJLp{,}@*) (*@\HLJLn{params}@*)(*@\HLJLp{,}@*) (*@\HLJLn{data}@*)(*@\HLJLp{,}@*) (*@\HLJLn{opt}@*)(*@\HLJLp{,}@*) (*@\HLJLn{cb}@*) (*@\HLJLoB{=}@*) (*@\HLJLn{cb}@*)(*@\HLJLp{)}@*)

(*@\HLJLnf{plot}@*)(*@\HLJLp{(}@*)(*@\HLJLnf{solve}@*)(*@\HLJLp{(}@*)(*@\HLJLnf{remake}@*)(*@\HLJLp{(}@*)(*@\HLJLn{prob}@*)(*@\HLJLp{,}@*)(*@\HLJLn{p}@*)(*@\HLJLoB{=}@*)(*@\HLJLn{Flux}@*)(*@\HLJLoB{.}@*)(*@\HLJLnf{data}@*)(*@\HLJLp{(}@*)(*@\HLJLn{p}@*)(*@\HLJLp{)),}@*)(*@\HLJLnf{SOSRI}@*)(*@\HLJLp{(),}@*)(*@\HLJLn{saveat}@*)(*@\HLJLoB{=}@*)(*@\HLJLnfB{0.1}@*)(*@\HLJLp{),}@*)(*@\HLJLn{ylim}@*)(*@\HLJLoB{=}@*)(*@\HLJLp{(}@*)(*@\HLJLni{0}@*)(*@\HLJLp{,}@*)(*@\HLJLni{6}@*)(*@\HLJLp{))}@*)
\end{lstlisting}

\includegraphics[width=\linewidth]{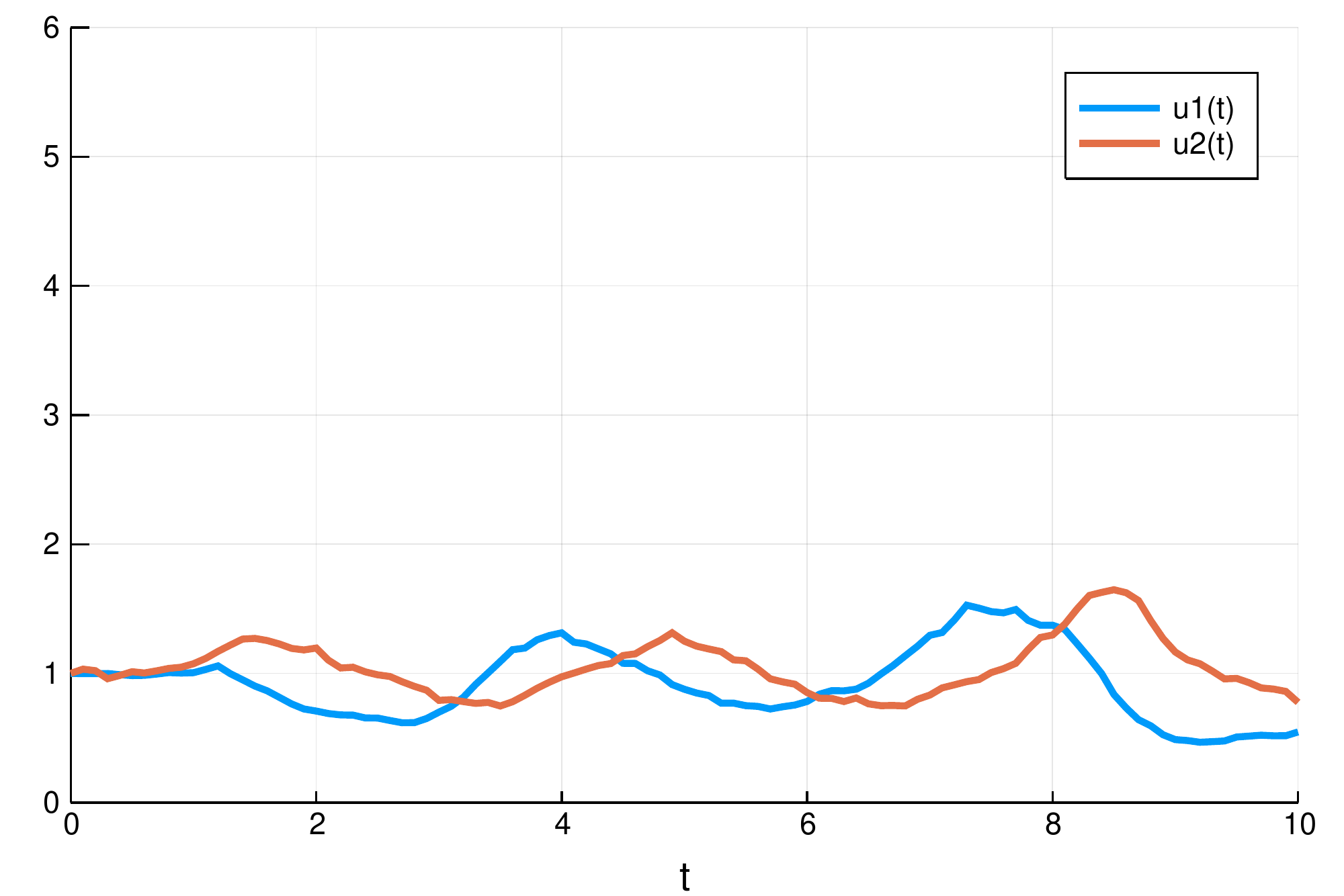}

(Animation omitted from the paper. Please see the \href{https://julialang.org/blog/2019/01/fluxdiffeq}{original blog post}).

\href{https://github.com/FluxML/model-zoo/blob/da4156b4a9fb0d5907dcb6e21d0e78c72b6122e0/other/diffeq/sde.jl}{This code can be found in the model-zoo}

And we can keep going. We can make neural versions of each of these. For example, \href{https://github.com/FluxML/model-zoo/blob/1d0447b6edb2b7fda04db663e745f347d7ae4ac6/other/diffeq/neural_sde.jl}{the model zoo contains an example training a neural SDE.} There are differential equations \href{http://docs.juliadiffeq.org/latest/tutorials/discrete_stochastic_example.html}{which are piecewise constant} used in biological simulations, or \href{http://docs.juliadiffeq.org/latest/tutorials/jump_diffusion.html}{jump diffusion equations from financial models}, and the solvers map right over to the Flux neural network framework through DiffEqFlux.jl. DiffEqFlux.jl uses only around 100 lines of code to pull this all off.

\section{Implementing the Neural ODE layer in Julia}
Let's go all the way back for a second and now implement the neural ODE layer in Julia. Remember that this is simply an ODE where the derivative function is defined by a neural network itself. To do this, let's first define the neural net for the derivative. In Flux, we can define a multilayer perceptron with 1 hidden layer and a \texttt{tanh} activation function like:

\begin{lstlisting}
(*@\HLJLn{dudt}@*) (*@\HLJLoB{=}@*) (*@\HLJLnf{Chain}@*)(*@\HLJLp{(}@*)(*@\HLJLnf{Dense}@*)(*@\HLJLp{(}@*)(*@\HLJLni{2}@*)(*@\HLJLp{,}@*)(*@\HLJLni{50}@*)(*@\HLJLp{,}@*)(*@\HLJLn{tanh}@*)(*@\HLJLp{),}@*)(*@\HLJLnf{Dense}@*)(*@\HLJLp{(}@*)(*@\HLJLni{50}@*)(*@\HLJLp{,}@*)(*@\HLJLni{2}@*)(*@\HLJLp{))}@*)
\end{lstlisting}

To define a \texttt{neural\_ode} layer, we then just need to give it a timespan and use the \texttt{neural\_ode} function:

\begin{lstlisting}
(*@\HLJLn{tspan}@*) (*@\HLJLoB{=}@*) (*@\HLJLp{(}@*)(*@\HLJLnfB{0.0f0}@*)(*@\HLJLp{,}@*)(*@\HLJLnfB{25.0f0}@*)(*@\HLJLp{)}@*)
(*@\HLJLn{x}@*)(*@\HLJLoB{->}@*)(*@\HLJLnf{neural{\_}ode}@*)(*@\HLJLp{(}@*)(*@\HLJLn{dudt}@*)(*@\HLJLp{,}@*)(*@\HLJLn{x}@*)(*@\HLJLp{,}@*)(*@\HLJLn{tspan}@*)(*@\HLJLp{,}@*)(*@\HLJLnf{Tsit5}@*)(*@\HLJLp{(),}@*)(*@\HLJLn{saveat}@*)(*@\HLJLoB{=}@*)(*@\HLJLnfB{0.1}@*)(*@\HLJLp{)}@*)
\end{lstlisting}

\begin{lstlisting}
#21 (generic function with 1 method)
\end{lstlisting}

As a side note, to run this on the GPU, it is sufficient to make the initial condition and neural network be on the GPU. This will cause the entire ODE solver's internal operations to take place on the GPU without extra data transfers in the integration scheme. This looks like:

\begin{lstlisting}
(*@\HLJLn{x}@*)(*@\HLJLoB{->}@*)(*@\HLJLnf{neural{\_}ode}@*)(*@\HLJLp{(}@*)(*@\HLJLnf{gpu}@*)(*@\HLJLp{(}@*)(*@\HLJLn{dudt}@*)(*@\HLJLp{),}@*)(*@\HLJLnf{gpu}@*)(*@\HLJLp{(}@*)(*@\HLJLn{x}@*)(*@\HLJLp{),}@*)(*@\HLJLn{tspan}@*)(*@\HLJLp{,}@*)(*@\HLJLnf{Tsit5}@*)(*@\HLJLp{(),}@*)(*@\HLJLn{saveat}@*)(*@\HLJLoB{=}@*)(*@\HLJLnfB{0.1}@*)(*@\HLJLp{)}@*)
\end{lstlisting}

\section{Understanding the Neural ODE layer behavior by example}
Now let's use the neural ODE layer in an example to find out what it means. First, let's generate a time series of an ODE at evenly spaced time points. We'll use the test equation from the Neural ODE paper.

\begin{lstlisting}
(*@\HLJLn{u0}@*) (*@\HLJLoB{=}@*) (*@\HLJLn{Float32}@*)(*@\HLJLp{[}@*)(*@\HLJLnfB{2.}@*)(*@\HLJLp{;}@*) (*@\HLJLnfB{0.}@*)(*@\HLJLp{]}@*)
(*@\HLJLn{datasize}@*) (*@\HLJLoB{=}@*) (*@\HLJLni{30}@*)
(*@\HLJLn{tspan}@*) (*@\HLJLoB{=}@*) (*@\HLJLp{(}@*)(*@\HLJLnfB{0.0f0}@*)(*@\HLJLp{,}@*)(*@\HLJLnfB{1.5f0}@*)(*@\HLJLp{)}@*)

(*@\HLJLk{function}@*) (*@\HLJLnf{trueODEfunc}@*)(*@\HLJLp{(}@*)(*@\HLJLn{du}@*)(*@\HLJLp{,}@*)(*@\HLJLn{u}@*)(*@\HLJLp{,}@*)(*@\HLJLn{p}@*)(*@\HLJLp{,}@*)(*@\HLJLn{t}@*)(*@\HLJLp{)}@*)
    (*@\HLJLn{true{\_}A}@*) (*@\HLJLoB{=}@*) (*@\HLJLp{[}@*)(*@\HLJLoB{-}@*)(*@\HLJLnfB{0.1}@*) (*@\HLJLnfB{2.0}@*)(*@\HLJLp{;}@*) (*@\HLJLoB{-}@*)(*@\HLJLnfB{2.0}@*) (*@\HLJLoB{-}@*)(*@\HLJLnfB{0.1}@*)(*@\HLJLp{]}@*)
    (*@\HLJLn{du}@*) (*@\HLJLoB{.=}@*) (*@\HLJLp{((}@*)(*@\HLJLn{u}@*)(*@\HLJLoB{.{\textasciicircum}}@*)(*@\HLJLni{3}@*)(*@\HLJLp{)}@*)(*@\HLJLoB{{\textquotesingle}}@*)(*@\HLJLn{true{\_}A}@*)(*@\HLJLp{)}@*)(*@\HLJLoB{{\textquotesingle}}@*)
(*@\HLJLk{end}@*)
(*@\HLJLn{t}@*) (*@\HLJLoB{=}@*) (*@\HLJLnf{range}@*)(*@\HLJLp{(}@*)(*@\HLJLn{tspan}@*)(*@\HLJLp{[}@*)(*@\HLJLni{1}@*)(*@\HLJLp{],}@*)(*@\HLJLn{tspan}@*)(*@\HLJLp{[}@*)(*@\HLJLni{2}@*)(*@\HLJLp{],}@*)(*@\HLJLn{length}@*)(*@\HLJLoB{=}@*)(*@\HLJLn{datasize}@*)(*@\HLJLp{)}@*)
(*@\HLJLn{prob}@*) (*@\HLJLoB{=}@*) (*@\HLJLnf{ODEProblem}@*)(*@\HLJLp{(}@*)(*@\HLJLn{trueODEfunc}@*)(*@\HLJLp{,}@*)(*@\HLJLn{u0}@*)(*@\HLJLp{,}@*)(*@\HLJLn{tspan}@*)(*@\HLJLp{)}@*)
(*@\HLJLn{ode{\_}data}@*) (*@\HLJLoB{=}@*) (*@\HLJLnf{Array}@*)(*@\HLJLp{(}@*)(*@\HLJLnf{solve}@*)(*@\HLJLp{(}@*)(*@\HLJLn{prob}@*)(*@\HLJLp{,}@*)(*@\HLJLnf{Tsit5}@*)(*@\HLJLp{(),}@*)(*@\HLJLn{saveat}@*)(*@\HLJLoB{=}@*)(*@\HLJLn{t}@*)(*@\HLJLp{))}@*)
\end{lstlisting}

Now let's pit a neural ODE against this data. To do so, we will define a single layer neural network which just has the same neural ODE as before (but lower the tolerances to help it converge closer, makes for a better animation!):

\begin{lstlisting}
(*@\HLJLk{using}@*) (*@\HLJLn{Random}@*)
(*@\HLJLn{Random}@*)(*@\HLJLoB{.}@*)(*@\HLJLnf{seed!}@*)(*@\HLJLp{(}@*)(*@\HLJLni{1}@*)(*@\HLJLp{)}@*) (*@\HLJLcs{{\#} Set a random seed so same starting parameters each time the}@*)
                (*@\HLJLcs{{\#} document is built!}@*)
(*@\HLJLn{dudt}@*) (*@\HLJLoB{=}@*) (*@\HLJLnf{Chain}@*)(*@\HLJLp{(}@*)(*@\HLJLn{x}@*) (*@\HLJLoB{->}@*) (*@\HLJLn{x}@*)(*@\HLJLoB{.{\textasciicircum}}@*)(*@\HLJLni{3}@*)(*@\HLJLp{,}@*)
             (*@\HLJLnf{Dense}@*)(*@\HLJLp{(}@*)(*@\HLJLni{2}@*)(*@\HLJLp{,}@*)(*@\HLJLni{50}@*)(*@\HLJLp{,}@*)(*@\HLJLn{tanh}@*)(*@\HLJLp{),}@*)
             (*@\HLJLnf{Dense}@*)(*@\HLJLp{(}@*)(*@\HLJLni{50}@*)(*@\HLJLp{,}@*)(*@\HLJLni{2}@*)(*@\HLJLp{))}@*)
(*@\HLJLn{ps}@*) (*@\HLJLoB{=}@*) (*@\HLJLn{Flux}@*)(*@\HLJLoB{.}@*)(*@\HLJLnf{params}@*)(*@\HLJLp{(}@*)(*@\HLJLn{dudt}@*)(*@\HLJLp{)}@*)
(*@\HLJLn{n{\_}ode}@*) (*@\HLJLoB{=}@*) (*@\HLJLn{x}@*)(*@\HLJLoB{->}@*)(*@\HLJLnf{neural{\_}ode}@*)(*@\HLJLp{(}@*)(*@\HLJLn{dudt}@*)(*@\HLJLp{,}@*)(*@\HLJLn{x}@*)(*@\HLJLp{,}@*)(*@\HLJLn{tspan}@*)(*@\HLJLp{,}@*)(*@\HLJLnf{Tsit5}@*)(*@\HLJLp{(),}@*)(*@\HLJLn{saveat}@*)(*@\HLJLoB{=}@*)(*@\HLJLn{t}@*)(*@\HLJLp{,}@*)(*@\HLJLn{reltol}@*)(*@\HLJLoB{=}@*)(*@\HLJLnfB{1e-7}@*)(*@\HLJLp{,}@*)(*@\HLJLn{abstol}@*)(*@\HLJLoB{=}@*)(*@\HLJLnfB{1e-9}@*)(*@\HLJLp{)}@*)
\end{lstlisting}

Notice that the \texttt{neural\_ode} has the same timespan and \texttt{saveat} as the solution that generated the data. This means that given an \texttt{x} (and initial value), it will generate a guess for what it thinks the time series will be where the dynamics (the structure) is predicted by the internal neural network. Let's see what time series it gives before we train the network. Since the ODE has two-dependent variables, we will simplify the plot by only showing the first. The code for the plot is:

\begin{lstlisting}
(*@\HLJLn{pred}@*) (*@\HLJLoB{=}@*) (*@\HLJLnf{n{\_}ode}@*)(*@\HLJLp{(}@*)(*@\HLJLn{u0}@*)(*@\HLJLp{)}@*) (*@\HLJLcs{{\#} Get the prediction using the correct initial condition}@*)
(*@\HLJLnf{scatter}@*)(*@\HLJLp{(}@*)(*@\HLJLn{t}@*)(*@\HLJLp{,}@*)(*@\HLJLn{ode{\_}data}@*)(*@\HLJLp{[}@*)(*@\HLJLni{1}@*)(*@\HLJLp{,}@*)(*@\HLJLoB{:}@*)(*@\HLJLp{],}@*)(*@\HLJLn{label}@*)(*@\HLJLoB{=}@*)(*@\HLJLs{"data"}@*)(*@\HLJLp{)}@*)
(*@\HLJLnf{scatter!}@*)(*@\HLJLp{(}@*)(*@\HLJLn{t}@*)(*@\HLJLp{,}@*)(*@\HLJLn{Flux}@*)(*@\HLJLoB{.}@*)(*@\HLJLnf{data}@*)(*@\HLJLp{(}@*)(*@\HLJLn{pred}@*)(*@\HLJLp{[}@*)(*@\HLJLni{1}@*)(*@\HLJLp{,}@*)(*@\HLJLoB{:}@*)(*@\HLJLp{]),}@*)(*@\HLJLn{label}@*)(*@\HLJLoB{=}@*)(*@\HLJLs{"prediction"}@*)(*@\HLJLp{)}@*)
\end{lstlisting}

\includegraphics[width=\linewidth]{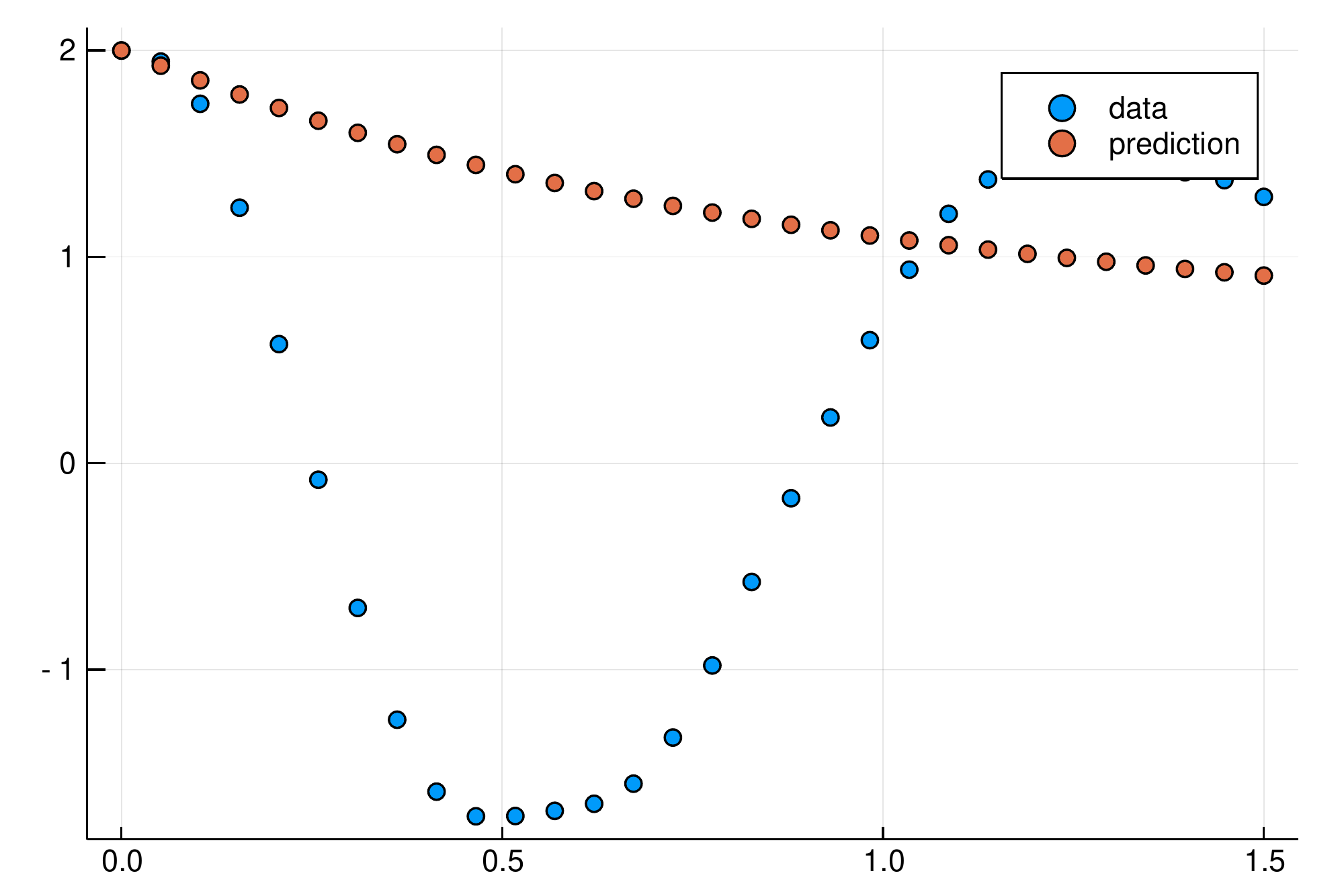}

But now let's train our neural network. To do so, define a prediction function like before, and then define a loss between our prediction and data:

\begin{lstlisting}
(*@\HLJLk{function}@*) (*@\HLJLnf{predict{\_}n{\_}ode}@*)(*@\HLJLp{()}@*)
  (*@\HLJLnf{n{\_}ode}@*)(*@\HLJLp{(}@*)(*@\HLJLn{u0}@*)(*@\HLJLp{)}@*)
(*@\HLJLk{end}@*)
(*@\HLJLnf{loss{\_}n{\_}ode}@*)(*@\HLJLp{()}@*) (*@\HLJLoB{=}@*) (*@\HLJLnf{sum}@*)(*@\HLJLp{(}@*)(*@\HLJLn{abs2}@*)(*@\HLJLp{,}@*)(*@\HLJLn{ode{\_}data}@*) (*@\HLJLoB{.-}@*) (*@\HLJLnf{predict{\_}n{\_}ode}@*)(*@\HLJLp{())}@*)
\end{lstlisting}

And now we train the neural network and watch as it learns how to predict our time series:

\begin{lstlisting}
(*@\HLJLn{data}@*) (*@\HLJLoB{=}@*) (*@\HLJLn{Iterators}@*)(*@\HLJLoB{.}@*)(*@\HLJLnf{repeated}@*)(*@\HLJLp{((),}@*) (*@\HLJLni{100}@*)(*@\HLJLp{)}@*)
(*@\HLJLn{opt}@*) (*@\HLJLoB{=}@*) (*@\HLJLnf{ADAM}@*)(*@\HLJLp{(}@*)(*@\HLJLnfB{0.1}@*)(*@\HLJLp{)}@*)
(*@\HLJLn{cb}@*) (*@\HLJLoB{=}@*) (*@\HLJLk{function}@*) (*@\HLJLp{()}@*) (*@\HLJLcs{{\#}callback function to observe training}@*)
  (*@\HLJLcs{{\#}display(loss{\_}n{\_}ode())}@*)
  (*@\HLJLcs{{\#} plot current prediction against data}@*)
  (*@\HLJLcs{{\#} Creates an animation}@*)
  (*@\HLJLcs{{\#}cur{\_}pred = Flux.data(predict{\_}n{\_}ode())}@*)
  (*@\HLJLcs{{\#}pl = scatter(t,ode{\_}data[1,:],label="data")}@*)
  (*@\HLJLcs{{\#}scatter!(pl,t,cur{\_}pred[1,:],label="prediction")}@*)
  (*@\HLJLcs{{\#}display(plot(pl))}@*)
(*@\HLJLk{end}@*)

(*@\HLJLcs{{\#} Display the ODE with the initial parameter values.}@*)
(*@\HLJLnf{cb}@*)(*@\HLJLp{()}@*)

(*@\HLJLn{Flux}@*)(*@\HLJLoB{.}@*)(*@\HLJLnf{train!}@*)(*@\HLJLp{(}@*)(*@\HLJLn{loss{\_}n{\_}ode}@*)(*@\HLJLp{,}@*) (*@\HLJLn{ps}@*)(*@\HLJLp{,}@*) (*@\HLJLn{data}@*)(*@\HLJLp{,}@*) (*@\HLJLn{opt}@*)(*@\HLJLp{,}@*) (*@\HLJLn{cb}@*) (*@\HLJLoB{=}@*) (*@\HLJLn{cb}@*)(*@\HLJLp{)}@*)

(*@\HLJLn{cur{\_}pred}@*) (*@\HLJLoB{=}@*) (*@\HLJLn{Flux}@*)(*@\HLJLoB{.}@*)(*@\HLJLnf{data}@*)(*@\HLJLp{(}@*)(*@\HLJLnf{predict{\_}n{\_}ode}@*)(*@\HLJLp{())}@*)
(*@\HLJLn{pl}@*) (*@\HLJLoB{=}@*) (*@\HLJLnf{scatter}@*)(*@\HLJLp{(}@*)(*@\HLJLn{t}@*)(*@\HLJLp{,}@*)(*@\HLJLn{ode{\_}data}@*)(*@\HLJLp{[}@*)(*@\HLJLni{1}@*)(*@\HLJLp{,}@*)(*@\HLJLoB{:}@*)(*@\HLJLp{],}@*)(*@\HLJLn{label}@*)(*@\HLJLoB{=}@*)(*@\HLJLs{"data"}@*)(*@\HLJLp{)}@*)
(*@\HLJLnf{scatter!}@*)(*@\HLJLp{(}@*)(*@\HLJLn{pl}@*)(*@\HLJLp{,}@*)(*@\HLJLn{t}@*)(*@\HLJLp{,}@*)(*@\HLJLn{cur{\_}pred}@*)(*@\HLJLp{[}@*)(*@\HLJLni{1}@*)(*@\HLJLp{,}@*)(*@\HLJLoB{:}@*)(*@\HLJLp{],}@*)(*@\HLJLn{label}@*)(*@\HLJLoB{=}@*)(*@\HLJLs{"prediction"}@*)(*@\HLJLp{)}@*)
\end{lstlisting}

\includegraphics[width=\linewidth]{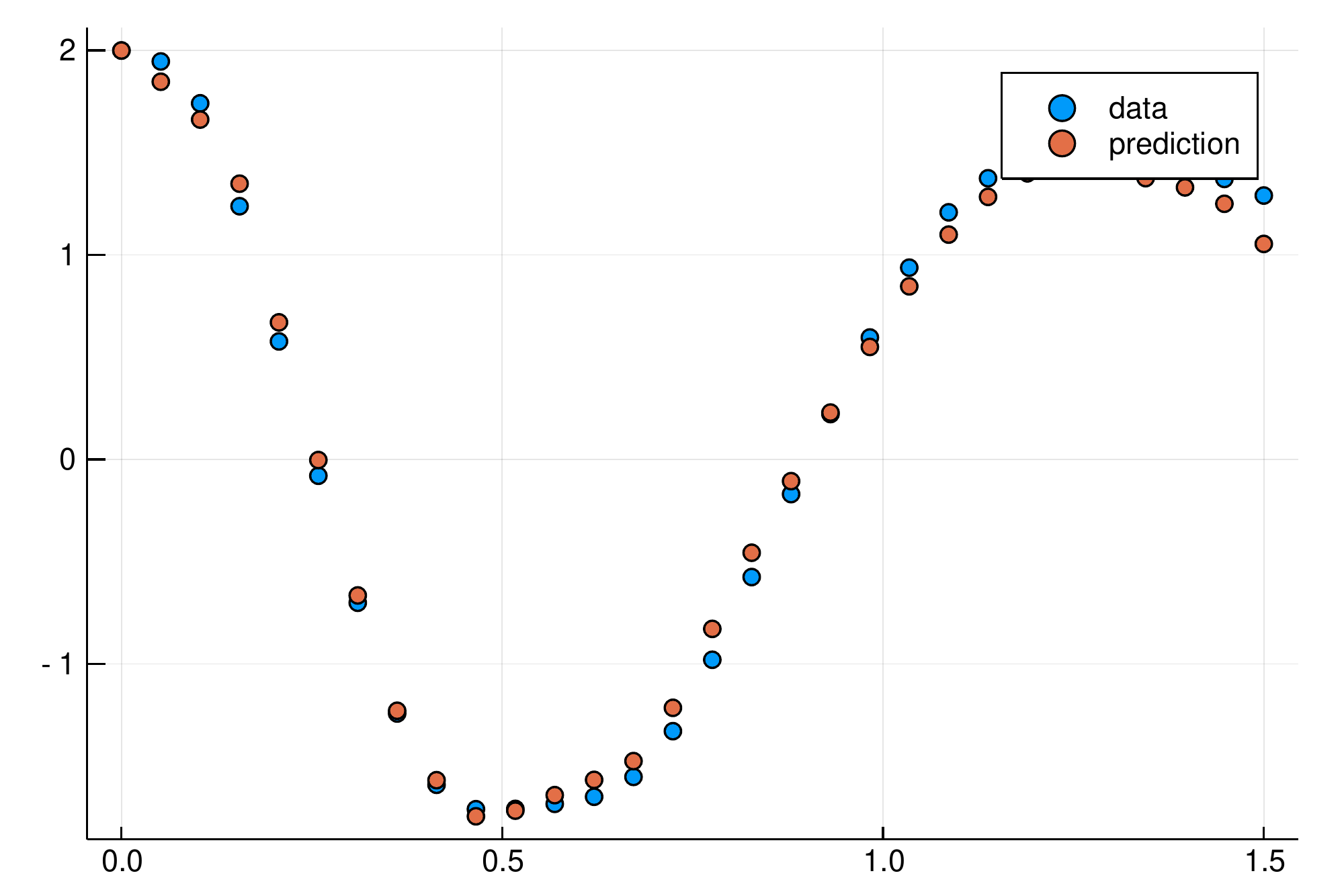}

(Animation omitted from the paper. Please see the \href{https://julialang.org/blog/2019/01/fluxdiffeq}{original blog post}).

\href{https://github.com/FluxML/model-zoo/blob/da4156b4a9fb0d5907dcb6e21d0e78c72b6122e0/other/diffeq/neural_ode.jl}{This code can be found in the model-zoo}

Notice that we are not learning a solution to the ODE. Instead, what we are learning is the tiny ODE system from which the ODE solution is generated. I.e., the neural network inside the neural\_ode layer learns this function:

Thus \textbf{it learned a compact representation of how the time series works}, and it can easily extrapolate to what would happen with different starting conditions. Not only that, it's a very flexible method for learning such representations. For example, if your data is unevenly spaced at time points \texttt{t}, just pass in \texttt{saveat=t} and the ODE solver takes care of it.

As you could probably guess by now, the DiffEqFlux.jl has all kinds of extra related goodies like Neural SDEs (\texttt{neural\_msde}) for you to explore in your applications.

\section{The core technical challenge: backpropagation through differential equation solvers}
Let's end by explaining the technical issue that needed a solution to make this all possible. The core to any neural network framework is the ability to backpropagate derivatives in order to calculate the gradient of the loss function with respect to the network's parameters. Thus if we stick an ODE solver as a layer in a neural network, we need to backpropagate through it.

There are multiple ways to do this. The most common is known as (adjoint) sensitivity analysis. Sensitivity analysis defines a new ODE whose solution gives the gradients to the cost function w.r.t. the parameters, and solves this secondary ODE. This is the method discussed in the neural ordinary differential equations paper, but actually dates back much further, and popular ODE solver frameworks like \href{http://people.cs.vt.edu/~asandu/Software/FATODE/index.html}{FATODE}, \href{https://web.casadi.org/}{CASADI}, and \href{https://computation.llnl.gov/projects/sundials/cvodes}{CVODES} have been available with this adjoint method for a long time (CVODES came out in 2005!). \href{http://docs.juliadiffeq.org/latest/analysis/sensitivity.html}{DifferentialEquations.jl has sensitivity analysis implemented too}

The efficiency problem with adjoint sensitivity analysis methods is that they require multiple forward solutions of the ODE. As you would expect, this is very costly. Methods like the checkpointing scheme in CVODES reduce the cost by saving closer time points to make the forward solutions shorter at the cost of using more memory. The method in the neural ordinary differential equations paper tries to eliminate the need for these forward solutions by doing a backwards solution of the ODE itself along with the adjoints. The issue with this is that this method implicitly makes the assumption that the ODE integrator is \href{https://www.physics.drexel.edu/~valliere/PHYS305/Diff_Eq_Integrators/time_reversal/}{reversible}. Sadly, there are no reversible adaptive integrators for first-order ODEs, so with no ODE solver method is this guaranteed to work. For example, here's a quick equation where a backwards solution to the ODE using the Adams method from the paper has >1700\% error in its final point, even with solver tolerances of 1e-12:

\begin{lstlisting}
(*@\HLJLk{using}@*) (*@\HLJLn{Sundials}@*)
(*@\HLJLk{function}@*) (*@\HLJLnf{lorenz}@*)(*@\HLJLp{(}@*)(*@\HLJLn{du}@*)(*@\HLJLp{,}@*)(*@\HLJLn{u}@*)(*@\HLJLp{,}@*)(*@\HLJLn{p}@*)(*@\HLJLp{,}@*)(*@\HLJLn{t}@*)(*@\HLJLp{)}@*)
 (*@\HLJLn{du}@*)(*@\HLJLp{[}@*)(*@\HLJLni{1}@*)(*@\HLJLp{]}@*) (*@\HLJLoB{=}@*) (*@\HLJLnfB{10.0}@*)(*@\HLJLoB{*}@*)(*@\HLJLp{(}@*)(*@\HLJLn{u}@*)(*@\HLJLp{[}@*)(*@\HLJLni{2}@*)(*@\HLJLp{]}@*)(*@\HLJLoB{-}@*)(*@\HLJLn{u}@*)(*@\HLJLp{[}@*)(*@\HLJLni{1}@*)(*@\HLJLp{])}@*)
 (*@\HLJLn{du}@*)(*@\HLJLp{[}@*)(*@\HLJLni{2}@*)(*@\HLJLp{]}@*) (*@\HLJLoB{=}@*) (*@\HLJLn{u}@*)(*@\HLJLp{[}@*)(*@\HLJLni{1}@*)(*@\HLJLp{]}@*)(*@\HLJLoB{*}@*)(*@\HLJLp{(}@*)(*@\HLJLnfB{28.0}@*)(*@\HLJLoB{-}@*)(*@\HLJLn{u}@*)(*@\HLJLp{[}@*)(*@\HLJLni{3}@*)(*@\HLJLp{])}@*) (*@\HLJLoB{-}@*) (*@\HLJLn{u}@*)(*@\HLJLp{[}@*)(*@\HLJLni{2}@*)(*@\HLJLp{]}@*)
 (*@\HLJLn{du}@*)(*@\HLJLp{[}@*)(*@\HLJLni{3}@*)(*@\HLJLp{]}@*) (*@\HLJLoB{=}@*) (*@\HLJLn{u}@*)(*@\HLJLp{[}@*)(*@\HLJLni{1}@*)(*@\HLJLp{]}@*)(*@\HLJLoB{*}@*)(*@\HLJLn{u}@*)(*@\HLJLp{[}@*)(*@\HLJLni{2}@*)(*@\HLJLp{]}@*) (*@\HLJLoB{-}@*) (*@\HLJLp{(}@*)(*@\HLJLni{8}@*)(*@\HLJLoB{/}@*)(*@\HLJLni{3}@*)(*@\HLJLp{)}@*)(*@\HLJLoB{*}@*)(*@\HLJLn{u}@*)(*@\HLJLp{[}@*)(*@\HLJLni{3}@*)(*@\HLJLp{]}@*)
(*@\HLJLk{end}@*)
(*@\HLJLn{u0}@*) (*@\HLJLoB{=}@*) (*@\HLJLp{[}@*)(*@\HLJLnfB{1.0}@*)(*@\HLJLp{;}@*)(*@\HLJLnfB{0.0}@*)(*@\HLJLp{;}@*)(*@\HLJLnfB{0.0}@*)(*@\HLJLp{]}@*)
(*@\HLJLn{tspan}@*) (*@\HLJLoB{=}@*) (*@\HLJLp{(}@*)(*@\HLJLnfB{0.0}@*)(*@\HLJLp{,}@*)(*@\HLJLnfB{100.0}@*)(*@\HLJLp{)}@*)
(*@\HLJLn{prob}@*) (*@\HLJLoB{=}@*) (*@\HLJLnf{ODEProblem}@*)(*@\HLJLp{(}@*)(*@\HLJLn{lorenz}@*)(*@\HLJLp{,}@*)(*@\HLJLn{u0}@*)(*@\HLJLp{,}@*)(*@\HLJLn{tspan}@*)(*@\HLJLp{)}@*)
(*@\HLJLn{sol}@*) (*@\HLJLoB{=}@*) (*@\HLJLnf{solve}@*)(*@\HLJLp{(}@*)(*@\HLJLn{prob}@*)(*@\HLJLp{,}@*)(*@\HLJLnf{CVODE{\_}Adams}@*)(*@\HLJLp{(),}@*)(*@\HLJLn{reltol}@*)(*@\HLJLoB{=}@*)(*@\HLJLnfB{1e-12}@*)(*@\HLJLp{,}@*)(*@\HLJLn{abstol}@*)(*@\HLJLoB{=}@*)(*@\HLJLnfB{1e-12}@*)(*@\HLJLp{)}@*)
(*@\HLJLn{prob2}@*) (*@\HLJLoB{=}@*) (*@\HLJLnf{ODEProblem}@*)(*@\HLJLp{(}@*)(*@\HLJLn{lorenz}@*)(*@\HLJLp{,}@*)(*@\HLJLn{sol}@*)(*@\HLJLp{[}@*)(*@\HLJLk{end}@*)(*@\HLJLp{],(}@*)(*@\HLJLnfB{100.0}@*)(*@\HLJLp{,}@*)(*@\HLJLnfB{0.0}@*)(*@\HLJLp{))}@*)
(*@\HLJLn{sol}@*) (*@\HLJLoB{=}@*) (*@\HLJLnf{solve}@*)(*@\HLJLp{(}@*)(*@\HLJLn{prob}@*)(*@\HLJLp{,}@*)(*@\HLJLnf{CVODE{\_}Adams}@*)(*@\HLJLp{(),}@*)(*@\HLJLn{reltol}@*)(*@\HLJLoB{=}@*)(*@\HLJLnfB{1e-12}@*)(*@\HLJLp{,}@*)(*@\HLJLn{abstol}@*)(*@\HLJLoB{=}@*)(*@\HLJLnfB{1e-12}@*)(*@\HLJLp{)}@*)
(*@\HLJLnd{@show}@*) (*@\HLJLn{sol}@*)(*@\HLJLp{[}@*)(*@\HLJLk{end}@*)(*@\HLJLp{]}@*)(*@\HLJLoB{-}@*)(*@\HLJLn{u0}@*)
\end{lstlisting}

\begin{lstlisting}
sol[end] - u0 = [-17.5445, -14.7706, 39.7985]
3-element Array{Float64,1}:
 -17.54451936669701 
 -14.770566993004616
  39.7984879277305
\end{lstlisting}

(Here we once again use the CVODE C++ solvers from SUNDIALS since they are a close match to the SciPy integrators used in the neural ODE paper.)

This inaccuracy is the reason why the method from the neural ODE paper is not implemented in software suites, but it once again highlights a detail. Not all ODEs will have a large error due to this issue. And for ODEs where it's not a problem, this will be the most efficient way to do adjoint sensitivity analysis. And this method only applies to ODEs. Not only that, it doesn't even apply to all ODEs. For example, ODEs with discontinuities (\href{http://docs.juliadiffeq.org/latest/features/callback_functions.html}{events}) are excluded by the assumptions of the derivation. Thus once again we arrive at the conclusion that one method is not enough.

In DifferentialEquations.jl have implemented many different methods for computing the derivatives of differential equations with respect to parameters. We have a \href{https://arxiv.org/abs/1812.01892}{recent preprint} detailing some of these results \cite{rackauckas_comparison_2018}. One of the things we have found is that direct use of automatic differentiation can be one of the most efficient and flexible methods. Julia's ForwardDiff.jl \cite{revels_forward-mode_2016}, Flux, and ReverseDiff.jl can directly be applied to perform automatic differentiation on the native Julia differential equation solvers themselves, and this can increase performance while giving new features. Our findings show that forward-mode automatic differentiation is fastest when there are less than 100 parameters in the differential equations, and that for >100 number of parameters adjoint sensitivity analysis is the most efficient. Even then, we have good reason to believe that \href{https://julialang.org/blog/2018/12/ml-language-compiler}{the next generation reverse-mode automatic differentiation via source-to-source AD, Zygote.jl} \cite{innes_dont_2018}, will be more efficient than all of the adjoint sensitivity implementations for large numbers of parameters.

Altogether, being able to switch between different gradient methods without changing the rest of your code is crucial for having a scalable, optimized, and maintainable framework for integrating differential equations and neural networks. And this is precisely what DiffEqFlux.jl gives the user direct access to. There are three functions with a similar API:

\begin{itemize}
\item \texttt{diffeq\_rd} uses Flux's reverse-mode AD through the differential equation solver.

\item \texttt{diffeq\_fd} uses ForwardDiff.jl's forward-mode AD through the differential equation solver.

\item \texttt{diffeq\_adjoint} uses adjoint sensitivity analysis to "backprop the ODE solver"

\end{itemize}
Therefore, to switch from a reverse-mode AD layer to a forward-mode AD layer, one simply has to change a single character. Since Julia-based automatic differentiation works on Julia code, the native Julia differential equation solvers will continue to benefit from advances in this field.

\section{Conclusion}
Machine learning and differential equations are destined to come together due to their complementary ways of describing a nonlinear world. In the Julia ecosystem we have merged the differential equation and deep learning packages in such a way that new independent developments in the two domains can directly be used together. We are only beginning to understand the possibilities that have opened up with this software. We hope that future blog posts will detail some of the cool applications which mix the two disciplines, such as embedding our coming pharmacometric simulation engine PuMaS.jl \cite{rackacukas_pumas:_2018} into the deep learning framework. With access to the full range of solvers for ODEs, SDEs, DAEs, DDEs, PDEs, discrete stochastic equations, and more, we are interested to see what kinds of next generation neural networks you will build with Julia.

\bibliographystyle{unsrt}
\bibliography{references}

\end{document}